\documentclass[10pt,journal]{IEEEtran}
\ifCLASSINFOpdf
\else
\fi

\usepackage{cite}
\usepackage{amsmath,amssymb,amsfonts}
\usepackage{algorithmic}
\usepackage{graphicx}
\usepackage{textcomp}
\usepackage{xcolor}
\usepackage{subcaption}
\usepackage{bbm}
\usepackage{booktabs}
\usepackage{multirow}
\usepackage{stfloats}
\usepackage{arydshln}
\usepackage{comment}

\begin{document}
% \title{Bare Demo of IEEEtran.cls\\ for IEEE Journals}
\title{Enabling On-Device CNN Training by Self-Supervised Instance Filtering and Error Map Pruning}

\author{Yawen Wu$^{\dagger}$, Zhepeng Wang$^{\dagger}$, Yiyu Shi$^{\ddagger}$, and Jingtong Hu$^{\dagger}$ \\
$^{\dagger}$Department of Electrical and Computer Engineering, University of Pittsburgh, USA \\
$^{\ddagger}$Department of Computer Science and Engineering, University of Notre Dame, USA \\
Email: yawen.wu@pitt.edu, zhepeng.wang@pitt.edu, yshi4@nd.edu, jthu@pitt.edu}

\markboth{}
{Shell \MakeLowercase{\textit{et al.}}: Bare Demo of IEEEtran.cls for IEEE Journals}



\maketitle

\begin{abstract}
This work aims to enable on-device training of convolutional neural networks (CNNs) by reducing the computation cost at training time.
CNN models are usually trained on high-performance computers and only the trained models are deployed to edge devices. 
But the statically trained model cannot adapt dynamically in a real environment and may result in low accuracy for new inputs.
On-device training by learning from the real-world data after deployment can greatly improve accuracy. 
However, the high computation cost makes training prohibitive for resource-constrained devices. 
To tackle this problem, we explore the computational redundancies in training and 
reduce the computation cost by two complementary approaches: 
self-supervised early instance filtering on data level and error map pruning on the algorithm level. 
The early instance filter selects important instances from the input stream to train the network and drops trivial ones. 
The error map pruning further prunes out insignificant computations when training with the selected instances.
Extensive experiments show that the computation cost is substantially reduced without any or with marginal accuracy loss. 
For example, when training ResNet-110 on CIFAR-10,
we achieve 68\% computation saving while preserving full accuracy and 75\% computation saving with a marginal accuracy loss of 1.3\%. Aggressive computation saving of 96\% is achieved with 
less than 0.1\%
accuracy loss when quantization is integrated into the proposed approaches.
Besides, when training LeNet on MNIST, 
we save 79\% computation while boosting accuracy by 0.2\%.
\end{abstract}


% \fi
\IEEEpeerreviewmaketitle

\vspace{-20pt}
\section{Introduction}

The maturation of deep learning has enabled on-device intelligence for Internet of Things (IoT) devices. Convolutional neural network (CNN), as an effective deep learning model, has been intensively deployed on IoT devices to extract information from the sensed data, such as smart cities \cite{song2018situ}, smart agriculture \cite{zhu2018deep} and wearable devices\cite{bhattacharya2016sparsification}. 
The models are initially trained on high-performance computers (HPCs) and then deployed to IoT devices for inference. However, in the physical world, the statically trained model cannot adapt to the real world dynamically and may result in low accuracy for new input instances. On-device training has the potential to learn from the environment and update the model in-situ. This enables incremental/lifelong learning \cite{ross2008incremental} to train an existing model to update its knowledge, and device personalization \cite{rudovic2018personalized} by learning features from the specific user and improving model accuracy. 
Federated learning \cite{federatedlearning} is another application scenario of on-device training, where a large number of devices (typically mobile phones) collaboratively learn a shared model while keeping the training data on personal devices to protect privacy. 
Since each device still computes the full model update by expensive training process, 
the computation cost of training needs to be greatly reduced to make federated learning realistic.

While the efficiency of training in HPCs can always be improved by allocating more computing resources, such as 1024 GPUs \cite{akiba2017extremely}, training on resource-constrained IoT devices remains prohibitive.
The main problem 
is the large gap between the high computation and energy demand of training and the limited computing resource and battery on IoT devices.
For example, training ResNet-110 \cite{he2016deep} on a 32x32 input image takes 780M FLOPs, which is prohibitive for IoT devices.
Besides, since computation directly translates into energy consumption and IoT devices are usually battery-constrained \cite{han2015deep}, the high computation demand of training will quickly drain the battery.
While existing works \cite{jayakodi2018trading,panda2016conditional,yawen2020intermittent} effectively reduce the computation cost of inference 
by assigning input instances to different classifiers according to the difficulty, the computation cost of training is not reduced.

To address this challenge, this work aims to enable on-device training by significantly reducing the computation cost of training while preserving the desired accuracy. Meanwhile, the proposed techniques can also be adopted to improve training efficiency on HPCs.
To achieve this goal, we investigate the computation cost of the entire training cycle, aiming to eliminate unnecessary computations while keeping full accuracy. We made the following two observations:
\emph{First}, not all the input instances are important for improving the model accuracy. Some instances are similar to the ones that the model has already been trained with and can be completely dropped to save computation.

Therefore, developing an approach to filter out unimportant instances can greatly reduce the computation cost.
\emph{Second}, for the important instances, not all the computation in the training cycle is necessary. Eliminating insignificant computations
will have a marginal influence on accuracy. 
In the backward pass of training, 
some channels in the error maps have small values. 
Pruning out these insignificant channels and corresponding computation will have a marginal influence on the final accuracy while saving a large portion of computation.

Based on the two observations, we propose a novel framework consisting of two complementary approaches to reduce the computation cost of training while preserving the full accuracy. 
The first approach is an early instance filter to select important instances from the input stream to train the network and drop trivial ones. The second approach is error map pruning to prune out insignificant computations in the backward pass when training with the selected instances.

In summary, the main contributions of this paper include:
\begin{itemize}
	\item \textbf{A framework to enable on-device training.} We propose a framework consisting of two approaches to eliminate unnecessary computation in training CNNs while preserving full network accuracy. 
	The first approach improves the training efficiency of both the forward and backward passes, and the second approach further reduces the computation cost in the backward pass.
	\item \textbf{Self-supervised early instance filtering (EIF) on the data level.} We propose an instance filter to predict the loss of each instance and develop a self-supervised algorithm to train the filter. Instances with predicted low loss are dropped before starting the training cycle to save computation. To train the filter simultaneously with the main network, we propose a self-supervised training algorithm including the adaptive threshold based labeling strategy, uncertainty sampling based instance selection algorithm, and weighted loss for biased high-loss ratio. 
	\item \textbf{Error map pruning (EMP) on the algorithm level.} We propose an algorithm to prune insignificant channels in error maps to reduce the computation cost in the backward pass. The channel selection strategy considers the importance of each channel on both the error propagation and the computation of the weight gradients to minimize the influence of pruning on the final accuracy.
\end{itemize}

We evaluate the proposed approaches on networks of different scales. ResNet and VGG are for on-device training of mobile devices, and LeNet is for tiny sensor node-level devices. Experimental results show the proposed approaches effectively reduce the computation cost of training without any or with marginal accuracy loss.

\vspace{-4pt}
\section{Background and Related Work}

\subsection{Background of CNN Training}
The training of CNNs is most commonly conducted with the mini-batch stochastic gradient descent (SGD) algorithm \cite{bottou2012stochastic}. It updates the model weights iteration-by-iteration using a mini-batch (e.g. 128) of input instances. 
For each instance in the mini-batch, a forward pass and a backward pass are conducted. The forward pass attempts to predict the correct outputs using current model weights. 
Then the backward pass back-propagates the loss through layers, which generates the error maps for each layer. 
Using the error maps, the gradient of the loss w.r.t. the model weights are computed. Finally, the model weights are updated by using the weight gradients and an optimization algorithm such as SGD. 

To provide labeled data for on-device training, labeling strategies from existing works can be used.
For example, the labels can come from aggregating inference results from neighbor devices \cite{lee2019neuro} (e.g. voting), employing spatial context information as the supervisory signals \cite{noroozi2016unsupervised, doersch2015unsupervised}, or naturally inferred from user interaction \cite{mcmahan2016communication,hard2018federated} such as  next-word-prediction in keyboard typing.
\vspace{-6pt}
\subsection{Related Work}
\textbf{Accelerated Training.}
There are a number of works on accelerating network training.
Stochastic depth\cite{huang2016deep} accelerates the training by randomly bypassing layers with the residual connection. E2Train\cite{wang2019e2} randomly drops mini-batches and selectively skips layers by using residual connections to save the computation cost. 
Different from \cite{wang2019e2}, which randomly drops mini-batches, we investigate the importance of each instance before keeping or dropping it. 
The input data from the real-word is not ideally shuffled and valuable instances for training can concentrate within one mini-batch. 
Simply dropping mini-batches can miss important instances for training the network.
Besides, the layer skipping in these two works rely on the ResNet architecture \cite{he2016deep}, and cannot be naturally extended to general CNNs. 
In contrast, our approaches are applicable to general CNNs.
OHEM \cite{shrivastava2016training} selects high-loss instances and drops low-loss ones to improve training efficiency. It computes the loss values of all instances in the forward pass and only keeps high-loss instances for the backward pass. The main drawback is that the computation in the forward pass of low-loss instances is wasted. Different from this, our approach predicts the loss of each instance and drops low-loss instances before starting the forward pass, which eliminates the computation cost of low-loss instances.

\textbf{Distributed Training.}
Another way to accelerate training is leveraging distributed training with abundant computing resources and large batch-sizes. 
\cite{akiba2017extremely} employs an extremely large batch size of 32K with 1024 GPUs to train ResNet-50 in 20 minutes. \cite{jia2018highly} integrates a mixed-precision method into distributed training and pushes the time to 6.6 minutes. 
However, these works target on leveraging highly-parallel computing resources to reduce the training time and actually increase the total computation cost, which is infeasible for training on resource-constrained IoT devices.

\textbf{Network Pruning during Training.}
Some works aim to train and prune the network architecture simultaneously. 
\cite{wen2016learning} proposes a pruning approach to sparsify weights during training. The goal is to generate a compact network for inference instead of improving training efficiency. In fact, it requires more time on training by first training the backbone then pruning it. Similarly, \cite{zhou2016less} prunes the sparse network during training to have a compact network for inference. 
However, these works only improve the inference efficiency and training computation cost is not reduced. 
\cite{alvarez2017compression, lym2019prunetrain} aim to accelerate training by reconfiguring the network to a smaller one during training.
The main drawback is that the network is pruned on the offline training dataset, and the ability of the pruned network for further on-device learning is compromised. 
Instead, we focus on reducing the computation cost of online training and the entire network architecture is preserved to keep the full ability for learning in an uncertain future.

\textbf{Network Compression and Neural Architecture Search.}
There are extensive explorations on network compression and neural architecture search (NAS).
\cite{li2016pruning,luo2017thinet} prune the network to generate a compact network for efficient inference. 
\cite{jiang2019accuracy,yang2020co,yang2020coe,jiang2020hardware}
search neural architectures for hardware-friendly inference.
\cite{lu2019neural} further considers quantization during NAS for efficient inference.
However, these works only aim to design network architectures for efficient inference. The computation cost of training is not considered.

\vspace{-6pt}
\section{Framework Overview}
\begin{figure}[!htb]
	\centering
	\vspace{-10pt}
	\includegraphics[width=0.9\columnwidth]{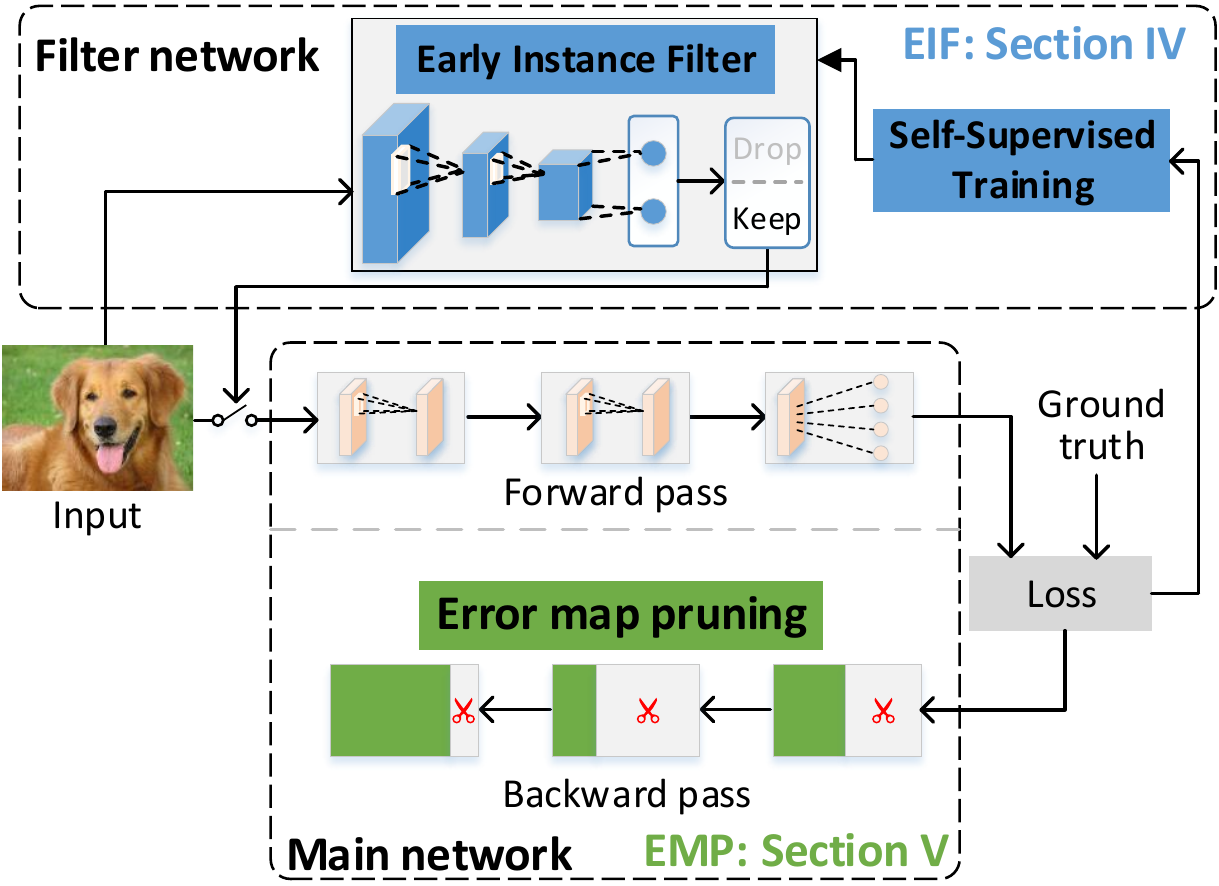}
	\vspace{-5pt}
% 	\captionsetup{labelfont={color=blue},font={color=blue}}
	\caption{Overview of early instance filtering (EIF) and error map pruning (EMP).}
	\label{fig:overview}
	\vspace{-14pt}
\end{figure}

The overview of the proposed framework is shown in Fig. \ref{fig:overview}. On top of the main neural network, a small instance filter network is proposed to select important instances from the input stream to train the network and drop trivial ones. When the input instances arrive, the early instance filter predicts the loss value for each instance as if the instance was fed into the main network and makes a binary decision to drop or preserve this instance. If the predicted loss is high and the instance is preserved, the main network will be invoked to start the forward and backward pass for training. Since the loss prediction is for the main network, once the main network is updated, the instance filter also needs to be trained for accurate loss prediction. The training of the instance filter is self-supervised based on the labeling strategy by the adaptive loss threshold, instance selection by uncertainty sampling, and the weighted loss for biased high-loss ratio, which will be introduced in Section \ref{sec:filter}.
Once important instances are selected, the error map pruning further reduces the computation cost of the backward pass. It prunes out channels in the error maps that have small contributions to the error propagation and gradient computation, which will be introduced in Section \ref{sec:pruning}. 

\vspace{-2pt}
\section{Self-Supervised Early Instance Filtering}\label{sec:filter}

\begin{figure}[!htb]
	\centering
	\vspace{-12pt}
	\includegraphics[width=1.0\columnwidth]{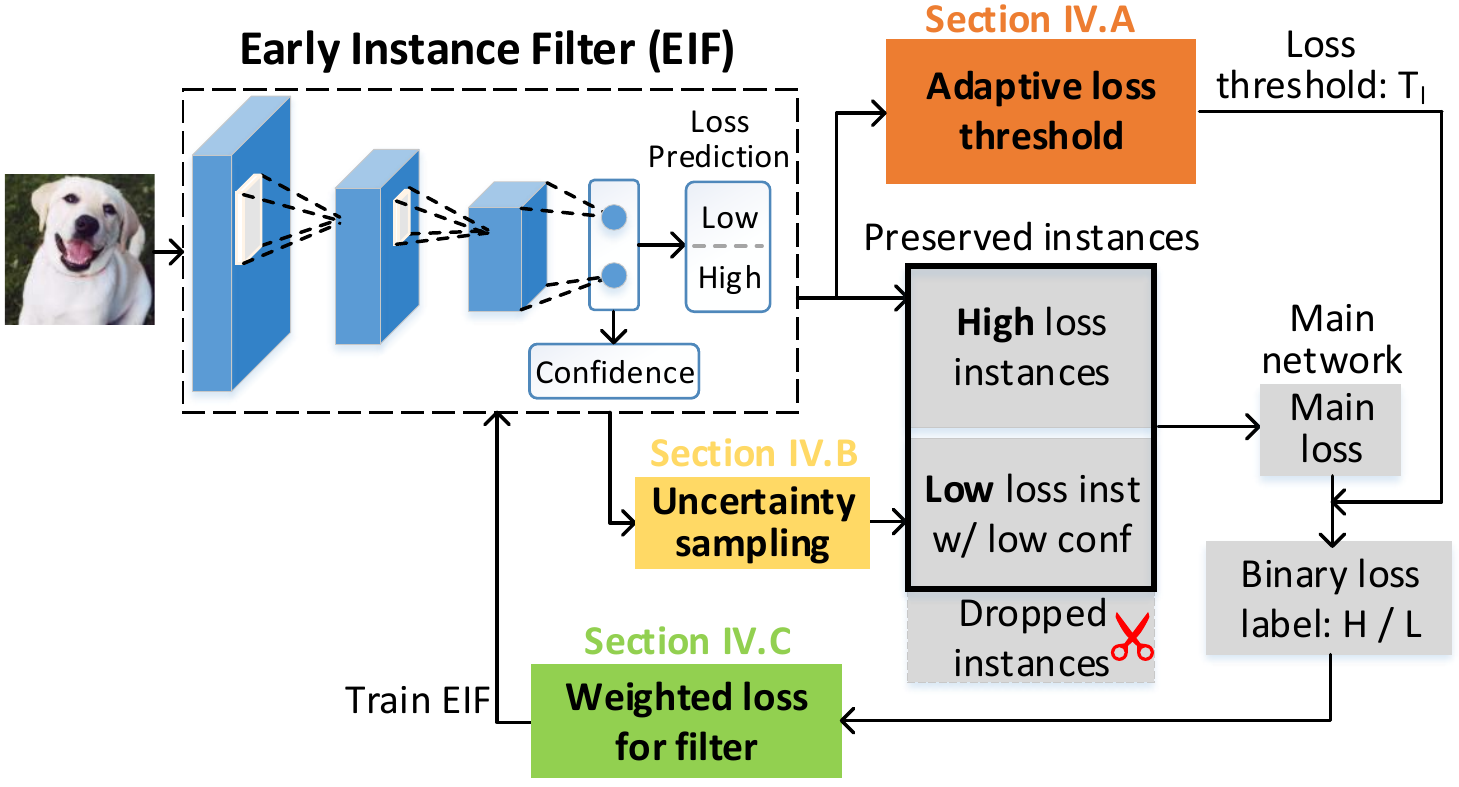}
	%	\vspace{-15pt}
	\vspace{-15pt}
% 	\captionsetup{labelfont={color=blue},font={color=blue}}
	\caption{Self-supervised training of early instance filter (EIF) by adaptive loss threshold, uncertainty sampling, and weighted loss.}
% 	\todo{switch B. C.}\todo{high loss ratio}}
	\label{fig:filter}
	\vspace{-10pt}
\end{figure}

The early instance filter (EIF) is used to select important instances for training the main network and drop trivial instances to reduce the computation cost of training. Since the main network is constantly being updated during training, it is essential to tune the EIF every time the main network is updated. 
In this way, the EIF can accurately select important instances based on the latest state of the main network.
In this section, we will first introduce the working flow of EIF to select instances for training the main model. Then we discuss the challenges for updating the EIF. After that, we present three approaches to address these challenges such that the EIF can be effectively updated.

To select important instances and drop trivial ones on-the-fly during training, the EIF predicts the loss value of each instance from the input stream without actually feeding the instance into the main network.
Trivial instances with predicted low loss are dropped before the forward pass, which eliminates the computation on the forward pass and more computationally intensive backward pass of the main network. Important instances with predicted high loss are preserved to complete the forward pass, calculate the loss, and finish the backward pass to compute the weight gradients to update the main network.
Kindly note that the instances are not pre-selected before the training starts. Instead, they are selected on-the-fly during training based on what the main network has and has not learned at the current state.

Fig. \ref{fig:filter} shows the working flow of EIF. The user first needs to pre-define a high-loss ratio $R_{set}$ (e.g. 10\%) such that these amounts of instances in the whole input stream will be predicted as high-loss and the others will be predicted as low-loss. 
Only instances predicted as high-loss will be used for training the main network.
When instances arrive sequentially,
the early instance filter predicts the loss value of each instance $i$ as binary high or low $y_{pred,i}=\{H,L\}$ for the main network such that the pre-defined high-loss ratio is satisfied. The filter also produces the confidence of each loss prediction, represented by the entropy of the loss prediction.
Since the loss prediction by the EIF network is for the main network and the main network is constantly being updated, it is essential to re-train the EIF network every time the main network is updated to realize accurate loss prediction. However, there are several challenges in realizing automatic self-supervised training for the EIF network. In this section, we will first present three major challenges. Then, we will present three techniques to address these challenges: adaptive loss threshold, uncertainty sampling, and weighted loss, as shown in Fig. \ref{fig:filter}.

\textbf{Challenges:} During on-device training, instances with predicted low-loss are dropped before feeding to the main network to compute the actual loss, and their true loss values are unknown. Thus, we can only know the true loss values of instances with predicted high-loss, which brings two challenges. 
The \textit{first} challenge is how to 
label instances as high-loss or low-loss for training the EIF according to the pre-defined high-loss ratio.
For example, if we could know the loss values of all instances, defining a loss threshold to separate 10\% instances with the highest loss values is simply sorting all the loss values and finding the value for separation. 
Since the loss values of dropped instances are unknown, defining a loss threshold remains a challenge.

The \textit{second} challenge is that the EIF network can choose what instances will be used to train itself, which is not possible for normal CNN training. 
As long as the EIF network is not 100\% accurate, it will make wrong predictions. 
To avoid punishment, instead of adjusting its own weights to make accurate loss predictions, the filter will learn a shortcut by predicting all the new input instances as low loss and drop them. Since the dropped instances will not be fed to the main network, the EIF network will never know the ground truth of the losses and thus it will not be punished for doing so. In this way, EIF will think it makes perfect predictions. Dropping all the new instances prevents further training of the filter and main network.

The \textit{third} challenge is how to correctly train the filter
when the number of high-loss and low-loss instances is extremely unbalanced in the input stream. This is different from normal training datasets such as CIFAR-10 and ImageNet, in which the number of instances in each class is balanced.
The unbalanced number of high-loss and low-loss instances makes the EIF network training ineffective. 
For example, when the pre-defined high-loss ratio is relatively low (e.g. 10\%), simply predicting all the instances as low-loss will produce high accuracy of 90\% on the filter, for which it believes as a good result. However, this prediction is useless since it does not find any important instance to train the main network.

We will present three techniques to address these challenges.

\vspace{-8pt}
\subsection{Adaptive Loss Threshold Based Labeling Strategy}
The adaptive loss threshold is used to provide the ground truth (labels) for training the EIF.
With the adaptive loss threshold, we can label the loss of instances as high-loss or low-loss to train the EIF.
During the training of EIF and the main network, $R_{set}$ percent of instances will be predicted as high-loss by the EIF.
The true loss values of instances predicted as high-loss can be obtained on the main network.
However, we do not know the true loss values of instances predicted as low-loss since they are dropped before feeding into the main network.
With only partial loss values, defining an exact loss threshold (e.g. sorting all loss values and finding the threshold) is challenging.
Therefore, we aim to approximate the threshold. To achieve this, we first define true high (TH) instances as the instances with predicted high loss by the filter and labeled as high-loss by the loss threshold.
We will monitor the number of TH instances in the last $n$ mini-batches. 
Then we calculate the percentage $R_{TH}$ as the number of TH instances in the preserved ones over all the instances in last $n$ mini-batches. By comparing the percentage $R_{TH}$ with the pre-defined percentage $R_{set}$, the loss threshold is adjusted to draw $R_{TH}$ to the pre-defined percentage $R_{set}$. 

Formally, with adaptive loss threshold $T_l$, instances are labeled as high-loss or low-loss as follows.
\setlength{\abovedisplayskip}{0pt}
\setlength{\belowdisplayskip}{0pt}
\setlength{\abovedisplayshortskip}{0pt}
\setlength{\belowdisplayshortskip}{0pt}
\begin{equation}\label{eq:loss_label}
y_{i} = 
\begin{cases}
H &\text{if $loss_{i}\ge T_l$}\\
L &\text{otherwise}
\end{cases}
% , \forall i \in \{1...m\}
\end{equation}
where $loss_{i}$ is the loss value of instance $i$ computed by the main network. $T_l$ is the adaptive loss threshold.

The true high (TH) loss instance ratio $R_{TH}$ by the filter is defined as follows.
\begin{equation}\label{eq:true_high_ratio}
R_{TH} = \frac{1}{mn}\sum_{i=1}^{mn}\mathbb{I}(y_{pred,i}=H) \mathbb{I}(y_i=H)
\end{equation}
$\mathbb{I}(x)$ is an indicator function which equals 1 if $x$ is true and 0 otherwise. $y_{pred,i}$ is the binary prediction by the filter for instance $i$, and $y_i$ is the loss label by Eq.(\ref{eq:loss_label}).
$m$ is the batch size, and $n$ is the number of mini-batches to monitor for one update of the loss threshold. 

Based on the computed $R_{TH}$ and pre-defined $R_{set}$, the loss threshold $T_l$ is adjusted to draw $R_{TH}$ to $R_{set}$. When $R_{TH}$ is larger than $R_{set}$, too many instances are labeled and predicted as high loss, which indicates $T_l$ is too small. Therefore, $T_l$ will be incremented by multiplying with a factor larger than 1. 
Similarly, when $R_{TH}$ is smaller than $R_{set}$, $T_l$ is too large and will be attenuated. The loss threshold $T_l$ is adjusted as:
\begin{equation}\label{eq:threshold_adjust}
T_l = 
\begin{cases}
\alpha_1 T_l &\text{if $R_{TH}\ge R_{set}$}\\
\alpha_2 T_l &\text{otherwise}
\end{cases}
\end{equation}
$\alpha_1$ and $\alpha_2$ are two hyper-parameters where $\alpha_1$ is larger than 1 and $\alpha_2$ is smaller than 1 to define the step size.

The computed $R_{TH}$ is essential to the self-supervised training of the EIF.
More specifically, $R_{TH}$ controls the loss threshold $T_l$ by Eq.(\ref{eq:threshold_adjust}), which further controls the instance labels $y_{i}$ by Eq.(\ref{eq:loss_label}) for training the instance filter.
With the labels $y_{i}$, the filter will be trained accordingly to predict high-loss instances. 
The number of instances with predicted high-loss by the filter and labeled as high-loss will be used to compute the new $R_{TH}$ by Eq.(\ref{eq:true_high_ratio}), which further adjusts $T_l$.
This process continues for each mini-batch, which forms the self-supervised training of the instance filter. Leveraging the self-supervision, the loss threshold $T_l$ will be properly adjusted and the instance filter will be well-trained to track the latest state of the main network. In this way, the true high-loss ratio $R_{TH}$ affected by both the filter and the loss threshold will be kept at the set ratio $R_{set}$. The filter will effectively select $R_{set}$ percent important instances for training the main network.

\vspace{-8pt}
\subsection{Instance Selection by Uncertainty Sampling}\label{sec:uncertainsampling}

The main reason for the second challenge is that if an instance is dropped, it will never be fed to the main network and the EIF network will never know the ground truth of the loss. 
In this way, the labels (i.e. high-loss or low-loss) of the dropped instances for training the EIF will be unknown, and the EIF cannot be correctly trained.
To address this problem, we keep some instances with predicted low loss, which would be dropped, to augment the preserved instances for training the filter. In this way, wrong loss predictions of the dropped instances will also punish the filter, which forces it to actually learn to find important instances.  
To decide which instances to keep and minimize the number of selected instances, we employ uncertainty sampling \cite{konyushkova2017learning}. The dropped instances that the filter are least confident about will be fed into the main network to compute the loss value. To measure the confidence of loss prediction by the filter, we use the entropy defined as:
\begin{equation}\label{equ:entropy}
entropy(i)=-\sum_{c\in\{H,L\}} p_{i,c}\log{p_{i,c}},\quad p_{i,c}=prob(y_{pred,i}=c)
\end{equation}
$p_{i,c}$ is the computed probability by the filter of being high-loss ($c=H$) or low-loss ($c=L$) for instance $i$.
The smaller the entropy, the more confident the filter is about the prediction. 

Based on the entropy, we select from the dropped instances where the entropy is above the entropy threshold to augment the preserved instances for training the filter. The set of selected instances is defined as:
\begin{equation}\label{equ:uncerntainty_sample}
\mathcal{I}=\{i\ |\ i\in \{TL,FL\},\ entropy(i) > entropy_T\}
\end{equation}
where $entropy_T$ is the entropy threshold. 

\vspace{-6pt}
\subsection{Weighed Loss for Biased High-Loss Ratio}\label{sec:weighted_loss}

To address the third challenge, we propose to use the weighted loss function to make the EIF training process fair in treating the high-loss instances when their ratio is low.
In this way, the EIF can be trained to make accurate loss predictions and select important instances for training the main network.

Traditionally, for datasets with balanced classes, we use the average loss of each instance as the loss function of a mini-batch for training. In our case, based on the binary loss label $y_{i}$ in Eq.(\ref{eq:loss_label}) and the binary loss prediction $y_{pred,i}$ by the filter, the loss function for instance $i$ is defined by cross-entropy as:
\begin{equation}\label{equ:loss_oneinstance}
L_i=-\sum_{c\in\{H,L\}} \mathbb{I}(y_i=c)\log{p_{i,c}}
\end{equation}
where $p_{i,c}$ defined in Eq.(\ref{equ:entropy}) is the computed probability of being a high or low loss for instance $i$ by the filter. 
$L_i$ measures how well the loss prediction approximates the true loss label and will be minimized during training.
The average loss will be the average loss value of each preserved instance in a mini-batch.
However, when the pre-defined high-loss ratio is not 50\% and makes the number of high-loss and low-loss instances unbalanced, 
directly using the average loss will result in effective training of the EIF.

To understand the inefficiency of training with the average loss, we define 
the weighted loss for preserved instances in a mini-batch to train the filter as:
\begin{equation}\label{equ:loss_filter}
L=\sum_{i\in TH}{w_{H}L_i} + \sum_{j\in FH}{w_{L}L_j} + \sum_{p\in TL}{w_{L}L_p} + \sum_{q\in FL}{w_{H}L_q}
\end{equation}
where $TH$, $FH$, $TL$ and $FL$ represent true high, false high, true low and false low loss instances, respectively. 
$TH$ and $FH$ are instances with predicted high loss and labeled as $H$ and $L$ by Eq.(\ref{eq:loss_label}), respectively.
$TL$ and $FL$ are instances with predicted low loss and selected by uncertainty sampling in Eq.(\ref{equ:uncerntainty_sample}), which have loss labels $L$ and $H$, respectively.
The weights 
$w_{H}$ and $w_{L}$ represent how important the true high loss (instances with loss label $H$, including $TH$ and $FL$) and true low loss (instances with loss label $L$, including $TL$ and $FH$) instances are, respectively.
$w_{H}$ and $w_{L}$ are normalized such that the weights of all instances in Eq.(\ref{equ:loss_filter}) sum up to 1.

When the pre-defined high-loss ratio $R_{set}$ is not 50\%, the number of high-loss and low-loss instances will be not equal in the input stream. This makes training the EIF with the average loss ineffective.
For example, when $R_{set}$ is set to $10\%$, only 10\% of the instances streamed in will be labeled as high-loss by the adaptive loss threshold. 
In this way, 90\% elements in Eq.(\ref{equ:loss_filter}) will be low-loss instances 
and dominate the loss. If we were using average loss, all the weights will be the same.
To minimize the loss when training the filter, simply predicting all instances as low-loss will produce small loss values on the dominating second and third elements in Eq.(\ref{equ:loss_filter}), and hence the total loss, which prevents effective training of the filter.

To address this problem, we make the weights biased by setting $w_H=\frac{1}{R_{set}}$ and $w_L=\frac{1}{1-R_{set}}$. In this way, we have $w_H\times percent(H=TH+FL)=w_L\times percent(L=TL+FH)$. The first and fourth sums in Eq.(\ref{equ:loss_filter}) correspond to the high-loss ($H$) instances. The second and third sums in Eq.(\ref{equ:loss_filter}) correspond to the low-loss ($L$) instances. By setting the weights in this way, the high-loss and low-loss instances will contribute equally to the total loss and will be treated fairly in training. In the above example, while the first and fourth sums only contribute to 10\% of the number of elements, the higher weight $w_H=\frac{1}{0.1}=10$ makes them equally important as the second and third sums, which have lower weight $w_H=\frac{1}{0.9}=1.1$. Therefore, the instance filter can be correctly trained with the unbalanced number of high-loss and low-loss instances and accurately predict high-loss ones.

With the predicted high-loss instances by the filter, the selected instances by uncertainty sampling, and the weighted loss function for training, the filter is effectively trained to predict high-loss instances for training the main network.

\vspace{-6pt}
\section{Error Map Pruning in Backward Pass}\label{sec:pruning}
\begin{figure}[!htb]
	\centering
	\vspace{-8pt}
	\includegraphics[width=0.9\columnwidth]{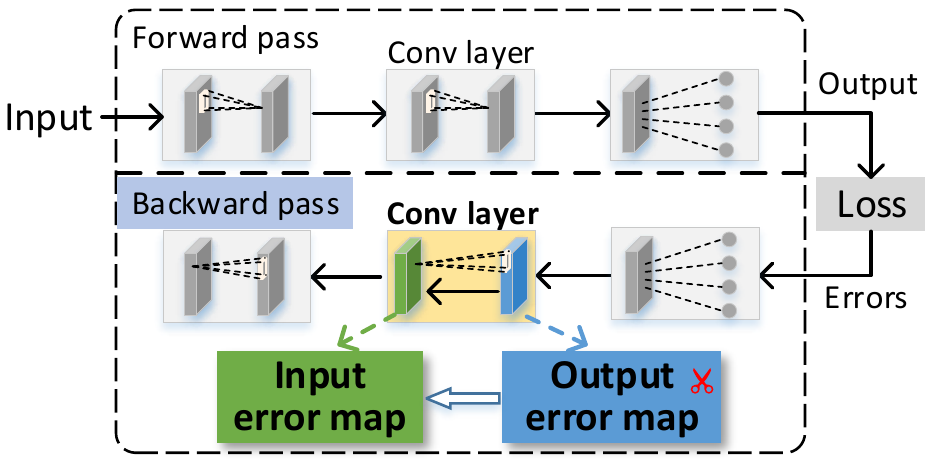}
	\vspace{-5pt}
	\caption{Error maps of convolutional layers in back-propagation.}
	\label{fig:error_map}
	\vspace{-10pt}
\end{figure}
When training with the selected instances, the computation in the backward pass can be further reduced by error map pruning (EMP). Since the backward pass takes about 2/3 computation cost of training, reducing its computation can effectively reduce the total cost. As shown in Fig. \ref{fig:error_map}, in the backward pass of training, the back-propagation propagates the errors layer-by-layer from the last layer to the first layer. We focus on pruning convolutional layers because they dominate the computation cost in the backward pass. Within one convolutional layer, the input error map is generated from the output error map of the same layer. The output error map consists of many channels. We aim to prune the insignificant channels to reduce the computation cost of training.

Given a pruning ratio, we need to keep the most representative channels in the error map to maintain as much information such that the training accuracy is retained. The main idea of the proposed channel selection strategy is to prune the channels that have the least influence on both error propagation and the computation of the weight gradients.

\vspace{-12pt}
\subsection{Channel Selection to Minimize Reconstruction Error on Error Propagation}
\begin{figure}[!htb]
	\centering
	\vspace{-19pt}
	\includegraphics[width=\columnwidth]{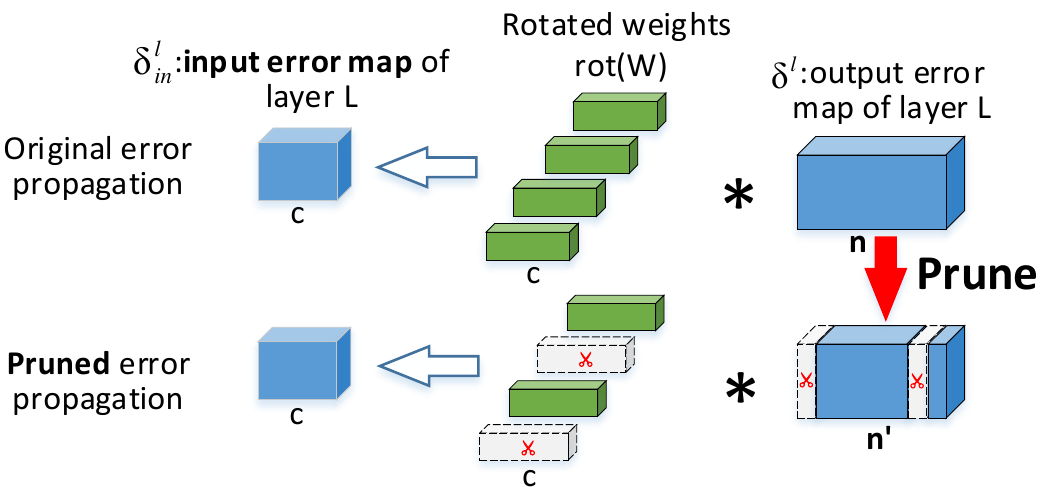}
	\vspace{-5pt}
	\caption{Back-propagation of errors with pruned error map.}
	\label{fig:error_pruning}
	\vspace{-8pt}
\end{figure}

The first criterion to select the channels to be pruned is to minimize reconstruction error on error propagation.
The error propagation for one convolutional layer is shown on the top of Fig. \ref{fig:error_pruning}. Within one layer, the error propagation starts from the output error map $\delta^l$ shown on the right, convolves $\delta^l$ with the rotated kernel weights $rot(W^l)$ and generates the input error map $\delta_{in}^l$ on the left. 
The error propagation with pruned $\delta^l$ is shown on the bottom of Fig. \ref{fig:error_pruning}. The number of channels in $\delta^l$ is pruned from $n$ to $n^\prime$. When computing $\delta_{in}^l$, the computations corresponding to the pruned channels, which are convolutional operations between $\delta^l$ and the rotated weights, are removed.
In order to maintain training accuracy, we want to keep the input error map $\delta_{in}^l$ before and after the pruning as same as possible. In another word, we want to minimize the reconstruction error on the input error map.

Formally, without channel pruning of $\delta^l$, $\delta_{in}^l$ is computed as follows.
\begin{equation}\label{equ:full_error_map}
	\delta_{in}^l = \sum_{j=1}^{n} rot(W_j^l) * \delta_j^l
\end{equation}
where $\delta_{in}^l$ is the input error map consisting of $c$ channels, each with shape $[W_{in},H_{in}]$. $rot(W_j^l)$ is the rotated weights of $j$th convolutional kernel with shape $[c,k_w,k_h]$. $\delta_j^l$ is the $j$th channel of the output error map with shape $[W,H]$.

Given a pruning ratio $\alpha$ and an output error map $\delta^l$, we aim to reduce the number of channels in $\delta^l$ from $n$ to $n^\prime$ such that $\alpha=n^\prime/n$.
To minimize the reconstruction error on $\delta_{in}^l$, the channel selection problem is formulated as follows.
\begin{align}
&\underset{\boldsymbol{\beta}}{\arg \min } \left\| \delta_{in}^l - \sum_{j=1}^{n} rot(W_j^l) * (\delta_j^l \beta_j) \right\|_{2} \label{equ:optimization} \\
& \quad \text{s.t.}\quad \|\boldsymbol{\beta}\|_{0} = n^\prime
\end{align}
where $\boldsymbol{\beta}$ is the error map selection strategy, represented as a binary vector of length $n$. $\beta_j$ is the $j$th entry of $\boldsymbol{\beta}$, and $\beta_j=0$ means the $j$th channel of $\delta_j^l$ is pruned.
The $\ell_{2}$-norm $\|\boldsymbol{x}\|_{2}=\sqrt{\Sigma x^2}$ measures the reconstruction error on $\delta_{in}^l$.  

However, directly solving the minimization problem is prohibitive. 
$\delta_{in}^l$ in the problem is computed by Eq.(\ref{equ:full_error_map}), which completes all the computation in error propagation and defeats the purpose of saving computation.
To select channels to prune before starting the actual error propagation, we define the importance score as an indication of how much each channel will influence the value of $\delta_{in}^l$ and prune the least important channels to minimize the reconstruction error on $\delta_{in}^l$.

\textbf{Importance Score.} In Eq.(\ref{equ:optimization}), when a channel $\delta_j^l$ is pruned, the computation error on $\delta_{in}^l$ is caused by the pruned $rot(W_j^l) * \delta_j^l$. 
As a fast and accurate estimation of the magnitude of $rot(W_j^l) * \delta_j^l$, 
we define the importance score of channel $j$ as follows.
\begin{equation}\label{equ:score}
s_j=\gamma_1 \left\|W_j^l\right\|_{1} + \gamma_2 \left\|\delta_j^l\right\|_{1}
\end{equation}
where $\|W_j^l\|_{1}$ is $\ell_{1}$-norm of convolutional kernel $j$, computed by
$\sum_{i=1}^{c}|W_{j,i}^l|$. Here we remove the rotation on $W_j^l$ since it does not change the $\ell_{1}$-norm. $\|\delta_j^l\|_{1}$ is $\ell_{1}$-norm of the channel $j$ in the output error map, computed by the sum of its absolute values $\sum_{x=1}^{W}\sum_{y=1}^{H}|\delta_{j,x,y}^l|$. $\gamma_1$ and $\gamma_2$ are two hyper-parameters to adjust the weight of each $\ell_{1}$-norm. 

The importance score $s_j$ gives an expectation of the magnitude that a channel $j$ in $\delta^l$ contributes to $\delta_{in}^l$. Channels with small magnitudes in $\delta^l$ and corresponding kernel weights $|W_{j}^l|$ tend to produce trivial values in the input error map $\delta_{in}^l$, which can be pruned while minimizing the influence on $\delta_{in}^l$.

\vspace{-8pt}
\subsection{Channel Selection to Minimize Reconstruction Error on Gradient Computation}
\begin{figure}[!htb]
	\centering
	\vspace{-16pt}
	\includegraphics[width=\columnwidth]{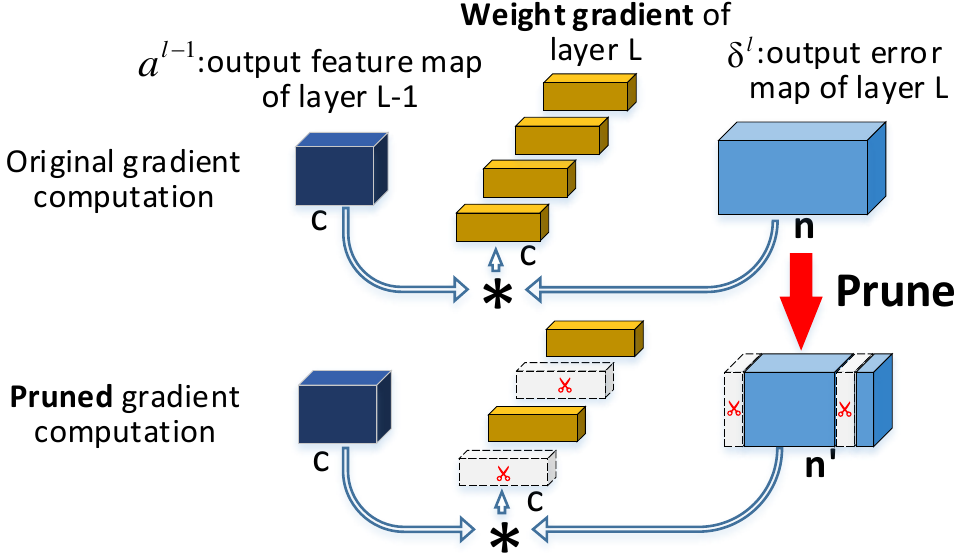}
	\vspace{-5pt}
	\caption{Computation of weight gradient with pruned error map.}
	\label{fig:gradient_pruning}
	\vspace{-10pt}
\end{figure}
The second criterion to select the channels to be pruned is to minimize the reconstruction error on the weight gradients.
The computation of the weight gradients without pruning is shown on the top of Fig. \ref{fig:gradient_pruning}. The output feature map $a^{l-1}$ of the previous layer convolves with one channel of the output error map $\delta^l$ to produce the gradient of one kernel. 
When some channels in $\delta^l$ are pruned, the computation of the weight gradients corresponding to the pruned channels is removed. To retain training accuracy, we want to keep the weight gradients before and after pruning as same as possible. 
Without channel pruning of $\delta^l$, the weight gradients of kernel $j$ are computed as follows.
\begin{equation}\label{equ:full_gradient}
	g_{w,j}^l = a^{l-1} * \delta_j^l,\quad \forall j\in \{1,...,n\}
\end{equation}
where $g_{w,j}^l$ is the weight gradients of kernel $j$ with shape $[c, k_w, k_h]$. $a^{l-1}$ is the output feature map of the previous layer $l-1$ with shape $[c, W_{in}, H_{in}]$. $\delta_j^l$ is the channel $j$ of the output error map in layer $l$, which has shape $[W, H]$. 

To determine the channel selection strategy $\boldsymbol{\beta}$ while minimizing the reconstruction error on the gradient computation, the channel selection problem is formulated as follows.
\begin{align}
& \underset{\boldsymbol{\beta}}{\arg \min } \sum_{j=1}^{n}\left\| g_{w,j}^l - a^{l-1} * (\delta_j^l \beta_j) \right\|_{2} \quad \text{s.t.}\ \|\boldsymbol{\beta}\|_{0} = n^\prime \label{equ:optimize_grad} 
\end{align}
Similar to Eq.(\ref{equ:optimization}), we use the $\ell_{2}$-norm $\|\cdot\|_{2}$ to measure the reconstruction error on the computation of the weight gradients for all the $n$ kernels incurred by the pruning. Similarly, solving this problem needs to complete all the gradient computation in Eq.(\ref{equ:full_gradient}) to get $g_{w,j}^l,j\in \{1,...,n\}$, which contradicts the goal to save computation. Thus, we define the importance score of each channel in $\delta^l$ for $g_{w}$ and prune the least important ones to minimize the reconstruction error on $g_{w}$.

\textbf{Importance Score.} In Eq.(\ref{equ:optimize_grad}), when a channel $\delta_j^l$ is pruned, the computation error is caused by the pruned $a^{l-1} * \delta_j^l$. Since $a^{l-1}$ is independent of $j$ and can be considered as a constant when measuring the importance of each channel $\delta_j^l$, we ignore $a^{l-1}$ and only include $\delta_j^l$ in the importance score of channel $j$, which is defined as follows.
\begin{equation}\label{equ:grad_score}
s_j=\left\|\delta_j^l\right\|_{1}
\end{equation}

\vspace{-6pt}
\subsection{Mini-Batch Pruning with Combined Importance Score}

To make the pruned channels for error propagation and gradient computation consistent with each other, we combine the importance score for these two processes. 
Then we scale it from instance-wise to batch-wise for mini-batch training.

The importance score $s_j$ for gradient computation in 
Eq.(\ref{equ:grad_score}) is a reduced form of Eq.(\ref{equ:score}) by setting $\gamma_1=0$ and $\gamma_2=1$. Therefore, we combine them into Eq.(\ref{equ:score}).
Based on the per-instance importance score of each channel, we can prune channels for a mini-batch of instances to reduce the computation while maintaining the accuracy. 
For a mini-batch of instances, we prune the same channels for all the instances.
The batch-wise importance score of one channel is calculated as 
$S_{j}=\sum_{i=1}^{m} s_j^i$.
$m$ is the batch size and $s_j^i$ is the importance score of channel $j$ for instance $i$.

With the batch-wise importance score, the error map pruning process for one convolutional layer is as follows.
Given a pruning ratio $\alpha$, $n(1-\alpha)$ channels in the output error map $\delta^l$ need to be pruned. First, for each channel $j$ in $\delta^l$, we calculate the batch-wise importance score $S_j$. Then the importance scores of all channels are sorted and $n(1-\alpha)$ channels with the smallest $S_j$ are marked as pruned.
Then the error propagation and the computation of the weight gradients corresponding to the pruned channels are skipped to save computation.

\textbf{Computation Reduction.}
With error map pruning, the computation cost of both the error propagation and the weight gradients is effectively reduced.
With pruning ratio $\alpha$, $1-\alpha$ computation in the error propagation and gradient computation is skipped, which saves about $1-\alpha$ computation in the backward pass of training.
More specifically, without pruning, for one instance the computation cost of error propagation for a convolutional layer $l$ in floating-point operations (FLOPs) is $FLOPs(\delta_{in}^l) = W_{in}H_{in}cnk_wk_h$.
When pruning the number of channels in $\delta^l$ from $n$ to $\alpha n$, the computation cost is reduced to $\alpha FLOPs(\delta_{in}^l)$.
For the computation of the weight gradients, before pruning the computation cost of $g_{w}^l$ is $FLOPs(g_{w}^l) = WHcnk_wk_h$.
With pruning ratio $\alpha$, the cost is reduced to $\alpha FLOPs(g_{w}^l)$.
In this way, $1-\alpha$ computation cost is reduced in the backward pass of convolutional layers. 

\textbf{Overhead Analysis.} The computation overhead of error map pruning is negligible. It is caused by the channel selection and the skipping of pruned channels. When using the $\ell_{1}$-norm strategy in Eq.(\ref{equ:score}) for the channel selection, the overhead is negligible because the sum over each kernel weight and each channel are relatively cheap compared with the expensive convolutional operation in the backward pass. For example, the channel selection of ResNet-110 consumes marginal
0.53\% FLOPs of the backward pass. For the overhead of skipping, since we employ structured pruning, skipping the pruned channels is simply skipping the computation involving the pruned channels, which has negligible overhead.

\vspace{-6pt}
\section{Experiments}\label{sec:exp}
We conduct extensive experiments to demonstrate the effectiveness of our approaches in terms of \emph{computation reduction}, \emph{energy saving}, \emph{accuracy}, \emph{convergence speed} and provide detailed \emph{analysis}.
The evaluation is on four network architectures and four datasets. We first evaluate EIF and then evaluate the combined EIF+EMP approach. After that, we evaluate the practical \emph{energy savings} on two edge devices.

\vspace{-12pt}
\subsection{Experimental setup}

\textbf{Datasets and Networks.}
We evaluate the proposed approaches on four datasets: CIFAR-10, CIFAR-100 \cite{cifar}, MNIST \cite{le2010mnist}, and ImageNet \cite{imagenet_cvpr09}. 
We use networks with different capacity to show the scalability of the proposed approaches. The networks include large-scale networks for mobile devices and small networks for tiny sensor nodes.
For large-scale networks, we employ two kinds of CNNs, including residual networks ResNet \cite{he2016deep} and plain networks VGG \cite{simonyan2014very}. ResNet-110, ResNet-74, and VGG-16 are evaluated on CIFAR-10/100. ResNet-18 and VGG-11 are evaluated on ImageNet.
For small networks, we use LeNet on MNIST. 

\textbf{Architectures of Instance Filter.}
We use different networks as the instance filter for different datasets. For CIFAR-10/100, we use ResNet-8. 
It has 7 convolutional layers and 1 fully-connected layer. 
The first layer is 3x3 convolutions with 16 filters. Then there is a stack of 3 residual blocks. Each block has 2 convolutional layers with kernel size 3x3. 
The numbers of filters in each block are \{16,32,64\}, respectively.
The network ends with a 10/100-way fully-connected layer.
For ImageNet, we use ResNet-10. 
It has 9 convolutional layers and 1 fully-connected layer. 
The first layer is 7x7 convolutions with 64 filters. Additional downsampling is conducted with a stride of 4 to reduce the computation cost. Then there is a stack of 4 residual blocks. Each block has 2 convolutional layers with kernel size 3x3. 
The numbers of filters in each block are \{64,128,256,512\}, respectively.
The network ends with a 1000-way fully-connected layer.
For MNIST, we use a slimmed LeNet with kernel size 3x3 and \{6,16\} filters for two convolutional layers, respectively.

The computation overhead of the EIF is negligible compared with the main networks.
For CIFAR-10/100, 
the computation required for the inference of the EIF is 
5.0\% of ResNet-110 and 4.1\% of VGG-16, respectively. 
The computation required for training the EIF is 
5.9\% of ResNet-110 and 4.8\% of VGG-16, respectively.
For ImageNet, the computation required for the inference and training of the EIF network is 3.4\% and 3.9\% of ResNet-18, and 0.81\% and 1.05\% of VGG-11, respectively.
For MNIST, the computation required for the inference and training of the EIF is 
9.5\% and 7.8\% of LeNet, respectively.

\textbf{Training Details.}
We train both the main network and the instance filter simultaneously \textit{from scratch}.
For ResNet-110, ResNet-74, and VGG-16, we employ the training settings in \cite{he2016deep}. We use SGD optimizer with momentum 0.9 and weight decay 0.0001 with batch size 128.
The models are trained for 64k iterations. The initial learning rate is 0.1 and decayed by a factor of 10 at 32k and 48k iterations.
For the instance filter, the learning rate is set to 0.1.
For ResNet-18 and VGG-11, similar training settings are employed except that the batch size is 256, the models are trained for 450k iterations, and the learning rate is decayed by a factor of 10 at 150k and 300k iterations.
For LeNet, the learning rate is 0.01 and the momentum is 0.5. The model is trained for 18.7k iterations with batch size 64. For the instance filter, the initial learning rate is 0.1 and decayed to 0.05 after 0.94k iterations. 

\textbf{Metrics.}
We evaluate the proposed approaches in two highly related but different metrics: the reduction of \emph{computation cost} and \emph{practical energy saving}.
The computation cost is measured in FLOPs, which is independent of the specific devices and a commonly used metric of computation cost\cite{sandler2018mobilenetv2}. The evaluation of the computation cost is conducted on NVIDIA P100 GPU with PyTorch 1.1 and measured by the THOP library \cite{thop}, which will be presented in Sections \ref{sec:exp_eif} - \ref{sec:exp_analysis}. 
The second metric, practical energy saving, depends on the devices and is measured on two edge devices (NVIDIA Jetson TX2 mobile GPU and MSP432 MCU), which will be presented in Section \ref{sec:hw_exp}.

\vspace{-8pt}
\subsection{Evaluating Early Instance Filtering (EIF)}\label{sec:exp_eif}
\vspace{-2pt}
\begin{figure}[!htb]
	\centering
	\vspace{-16pt}
	\includegraphics[width=0.9\columnwidth]{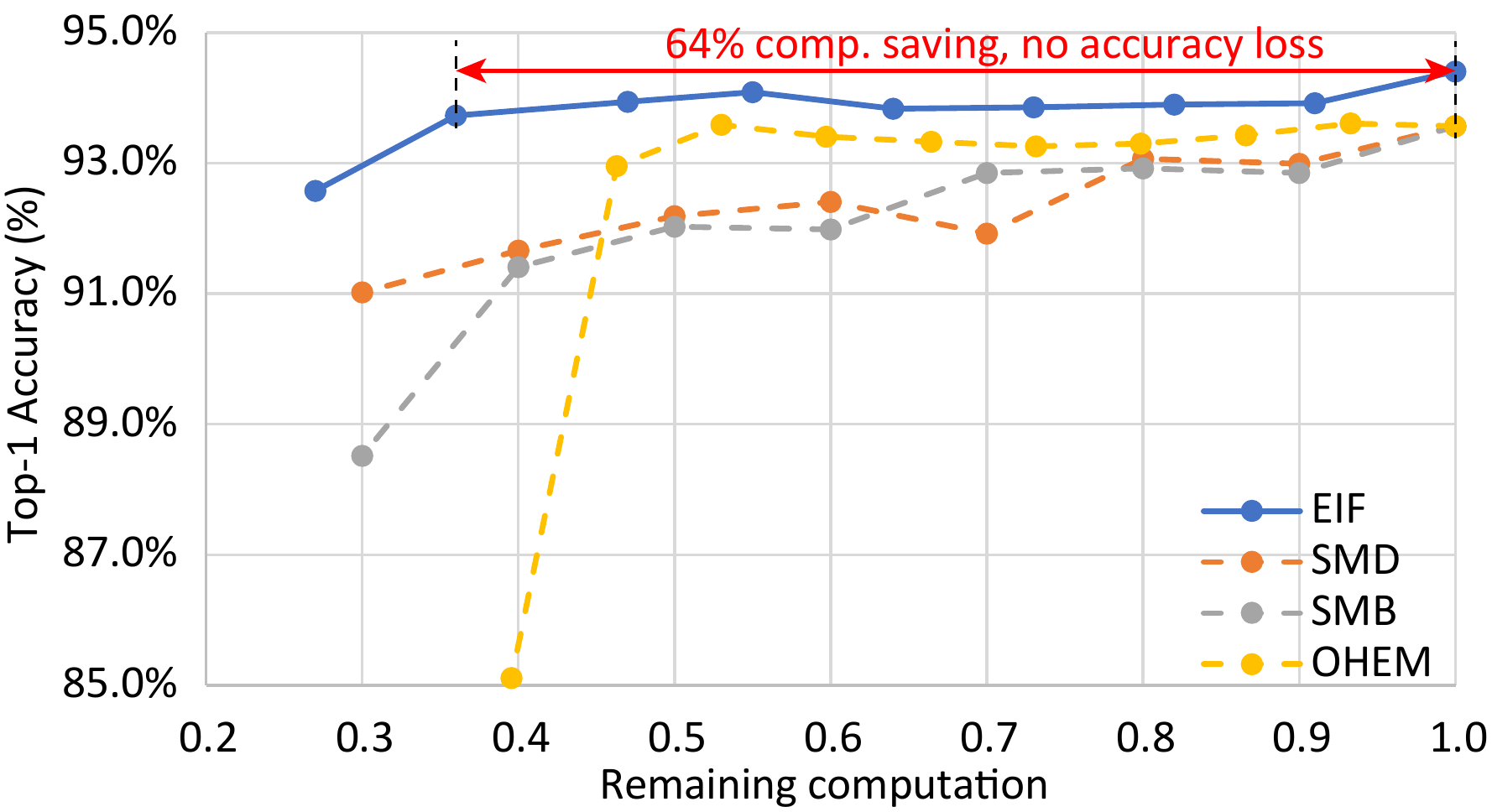}
	\vspace{-7pt}
	\caption{Top-1 accuracy by early instance filter (EIF) and baselines with ResNet-110 on CIFAR-10.}
	\vspace{-12pt}
	\label{fig:eif_res110}
% 	\vspace{-8pt}
\end{figure}
To show that the proposed early instance filtering (EIF) can effectively reduce the computation cost while maintaining or even boosting the accuracy, we compare with two start-of-the-art (SOTA) baselines and a standard training approach. 
Online hard example mining (OHEM) \cite{shrivastava2016training} selects hard examples for training by computing the loss values.
Stochastic mini-batch dropping (SMD) \cite{wang2019e2} is a SOTA on-device training approach, which randomly skips every mini-batch with a probability. 
SMB is the standard mini-batch training method by stochastic gradient descent (SGD), and the computation cost is adjusted by reducing the number of training iterations.

\textbf{Computation Reduction while Boosting Accuracy.}
The proposed EIF substantially outperforms the baselines in terms of both accuracy and computation reduction. As shown in Fig. \ref{fig:eif_res110}, when training ResNet-110 on CIFAR-10, with different remaining computation ratio, EIF consistently outperforms the baselines by a large margin. 
Compared with the full accuracy by SGD (e.g. SMB with remaining computation ratio 1.0), 
when using only 36.50\% remaining computation, EIF boosts the accuracy by 0.16\% (93.73\% vs. 93.57\%).
With only 55.45\% computation, EIF boosts the accuracy by 0.52\% (94.09\% vs. 93.57\%).
Compared with SMB and SMD, under different computation ratio, EIF achieves consistently higher accuracy with range [0.84\%, 2.32\%] and [0.83\%, 2.28\%], respectively.
The significant improvement is achieved because EIF selects instances by predicting the true loss value, instead of randomly dropping the instances.
Compared with OHEM, EIF consistently achieves higher accuracy with range [0.31\%, 0.98\%] under different computation ratio.
The improved accuracy and reduced computation cost show that the proposed instance filter effectively selects important instances for training to save computation cost.

\begin{figure*}[!htb]
	\centering
	%	\vspace{-15pt}
	\includegraphics[width=1.0\textwidth]{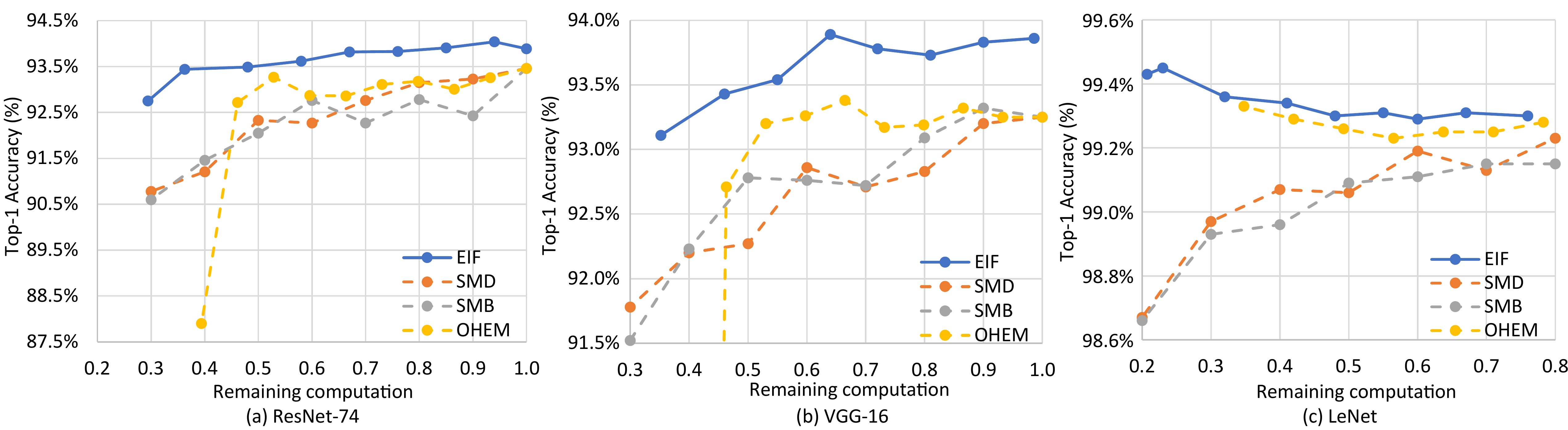}
	\vspace{-18pt}
	\caption{Top-1 accuracy by EIF and baselines with ResNet-74 and VGG-16 on CIFAR-10 and LeNet on MNIST.}
	\vspace{-15pt}
	\label{fig:eif_res74_vgg16_lenet}
% 	\vspace{-5pt}
\end{figure*}

To further evaluate EIF, we conduct experiments on training ResNet-74, VGG-16 and LeNet. Consistent accuracy improvement over the SOTA baselines is observed as shown in Fig. \ref{fig:eif_res74_vgg16_lenet}(a)(b)(c). 
For ResNet-74, with 63.74\% computation reduction, EIF achieves 0.02\% accuracy loss, while OHEM has a larger accuracy loss of 5.56\% with less computation reduction of 60.59\%. SMD and SMB achieve an accuracy loss of 2.25\% and 1.98\% with less computation reduction of 60\%.
Similar results are observed on VGG-16. 
For LeNet, EIF boosts the accuracy by 0.21\% (99.45\% over 99.24\%) with computation reduction 77.19\%. 

\vspace{-8pt}
\subsection{Evaluating EIF + EMP}
\begin{figure}[!htb]
	\centering
	\vspace{-6pt}
	\includegraphics[width=1.0\columnwidth]{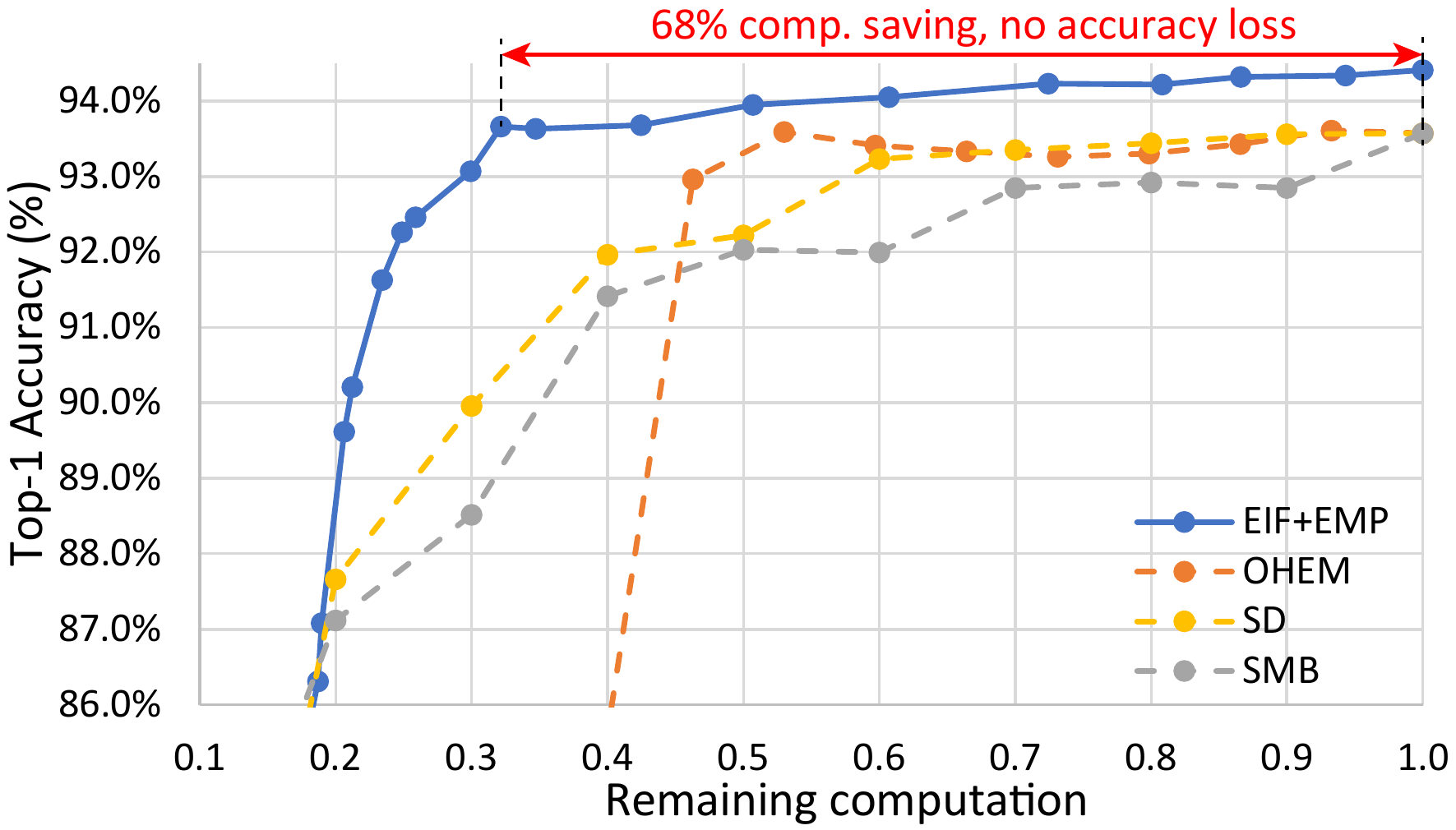}
	\vspace{-16pt}
	\caption{Accuracy of ResNet-110 on CIFAR-10 by EIF+EMP and baselines under different remaining computation ratios.}
	\vspace{-10pt}
	\label{fig:EIFEMP_res110}
% 	\vspace{-8pt}
\end{figure}

We evaluate the proposed framework EIF+EMP consisting of EIF and EMP and compare it with the SOTA baselines. 
Our approach effectively reduces the computation cost and achieves significantly better accuracy that SOTA baselines. Fig. \ref{fig:EIFEMP_res110} shows the accuracy of ResNet-110 on CIFAR-10 when trained by EIF+EMP and the baselines under different remaining computation ratio. 
Compared with EIF or EMP only, EIF+EMP achieves more aggressive computation reduction while preserving and even boosting the accuracy.
With EIF only, we achieve 63.50\% computation reduction without accuracy loss. With EMP only, we achieve 35.56\% computation reduction in the backward pass without accuracy loss and 62.22\% computation reduction with a slight accuracy loss of 0.72\%.
By the combined EIF+EMP, with up to 67.84\% computation reduction, we achieve no accuracy loss and boosts the accuracy by up to 0.84\% (94.41\% vs. 93.57\%). 

\begin{table}
	%	\vspace{-20pt}
	\centering
	\caption{Top-1 accuracy by EIF+EMP with ResNet-110, ResNet-74, VGG-16 on CIFAR-10 and LeNet on MNIST. 
	}
	\vspace{-5pt}
	\begin{tabular}{llll}
		\toprule 
		{Network}& \text {Method} & \text{Comp. Reduce} & \text {Accuracy}\\ \hline
		\multirow{6}{*}{ResNet-110} & \text {SGD(original)} & - & 93.57\% \\ 
		& \text {\textbf{EIF+EMP}} & \textbf{67.84}\% & \textbf{93.66}\% \\
		& \text {OHEM\cite{shrivastava2016training}} & 60.45\% & 85.11\% \\
		& \text {SD\cite{huang2016deep}} & 60.00\% & 91.96\% \\ \cline{2-4}
		& \text {\textbf{EIF+EMP+Q}} & \textbf{95.71\%} & \textbf{93.54\%} \\
		& \text {E2Train(+Q)\cite{wang2019e2}} & 90.13\% & 91.68\% \\
		\hline
		\multirow{6}{*}{ResNet-74} & \text {SGD(original)} & - & 93.46\% \\
		& \text {\textbf{EIF+EMP}} & \textbf{63.91}\% & \textbf{93.48}\% \\
		& \text {OHEM} & 60.59\% & 87.90\% \\
		& \text {SD} & 60.00\% & 90.99\% \\  \cline{2-4}
		& \text {\textbf{EIF+EMP+Q}} & \textbf{95.41\%} & \textbf{93.00\%} \\ 
		& \text {E2Train(+Q)} & 90.13\%  & 91.36\%  \\
		\hline
		\multirow{5}{*}{VGG-16} & \text {SGD(original) } & - & 93.25\% \\
		& \text {\textbf{EIF+EMP}} & \textbf{67.33\%} & \textbf{93.15\%} \\
		& \text {OHEM} & 60.41\% & 71.81\% \\
		\cline{2-4}
		& \text {\textbf{EIF+EMP+Q}} & \textbf{95.54\%} & \textbf{92.69\%} \\
		& \text {E2Train(+Q)} & - & - \\
		\hline
		\multirow{3}{*}{LeNet} & \text {SGD(original) } & - & 99.23\% \\
		& \text {\textbf{EIF+EMP}} & \textbf{78.60}\% & \textbf{99.47}\% \\
		& \text {OHEM} & 65.24\% & 99.33\% \\
		\bottomrule
	\end{tabular}
	\label{tbl:res110_quant}
	\vspace{-12pt}
\end{table}
We further evaluate EIF+EMP with more network architectures and datasets. 
Our approach substantially outperforms the baselines in terms of computation reduction and accuracy. 
We evaluate our approach with ResNet-110, ResNet-74 and VGG-16 on CIFAR-10 and LeNet on MNIST. 
For a fair comparison with E2Train \cite{wang2019e2}, which employs quantization\cite{banner2018scalable}, we use the same quantization scheme. When comparing with other baselines, we do not use quantization.
The experimental results are shown in Table \ref{tbl:res110_quant}. 
When training ResNet-74, our approach achieves 63.91\% computation saving without accuracy loss.
With quantization, our approach achieves 95.41\% computation saving with a marginal accuracy loss of 0.46\%. 
E2Train achieves less computation saving of 90.13\% and a much higher accuracy loss of 2.10\%. 
Similar results are observed on ResNet-110, VGG-16 and LeNet. 
SD and E2Train rely on the residual connections in ResNet and cannot be applied to VGG-16 and LeNet.
This result shows the proposed framework EIF+EMP achieves superior computation saving and significantly higher accuracy than baselines on different networks.

\begin{table}
	%	\vspace{-20pt}
	\centering
	\caption{Top-1 accuracy by EIF+EMP and baselines with ResNet-110 and VGG-16 on CIFAR-100.}
	\vspace{-6pt}
	\begin{tabular}{llll}
		\toprule 
		Network & Method & Comp. Reduce & Accuracy \\ \hline
		\multirow{8}{*}{ResNet-110} & SGD (Original) & - & 71.60\% \\
		& \multirow{2}{*}{\textbf{EIF+EMP}} & \textbf{50.02\%} & \textbf{72.02\%} \\
		&  & \textbf{56.24\%} & \textbf{71.63\%} \\
		& OHEM & 47.01\% & 69.98\% \\
		& SD & 50.00\% & 70.44\% \\
		& SMB & 50.00\% & 67.28\% \\ \cline{2-4}
		& \textbf{EIF+EMP+Q} & \textbf{92.92\%} & \textbf{71.29\%} \\
		& E2Train(+Q) & 90.13\% & 67.94\% \\
		\hline
		\multirow{5}{*}{VGG-16} & SGD (Original) & - & 71.56\% \\
		& \multirow{2}{*}{\textbf{EIF+EMP}} & \textbf{50.49\%} & \textbf{71.59\%} \\
		&  & \textbf{53.86\%} & \textbf{70.92\%} \\
		& OHEM & 46.99\% & 65.17\% \\
		& SMB & 50.00\% & 68.76\% \\ 
		\bottomrule
	\end{tabular}
	\vspace{-6pt}
	\label{tbl:eifemp_cifar100}
\end{table}

\textbf{Experiments on CIFAR-100.}
We further evaluate the proposed approaches on CIFAR-100 with ResNet-110 and VGG-16. EIF+EMP substantially outperforms the baselines in both computation reduction and accuracy. As shown in Table \ref{tbl:eifemp_cifar100}, with ResNet-110, EIF+EMP achieves 56.24\% computation reduction while preserving the full network accuracy, and 50.02\% computation reduction while boosting the accuracy by 0.42\%. The baselines OHEM, SD and SMB achieve much lower accuracy even with less computation reduction. 
With quantization, EIF+EMP achieves 92.92\% computation reduction with a marginal accuracy loss of 0.31\%, while E2Train has a much larger accuracy loss of 3.66\% with less computation reduction of 90.13\%.
For VGG-16, EIF+EMP achieves 50.49\% computation reduction without accuracy loss and 53.86\% computation reduction with only 0.64\% accuracy loss. The baselines OHEM and SMB achieve much larger accuracy loss of 6.39\% and 2.80\% with less computation reduction. SD and E2Train cannot be applied to VGG-16, which does not have residual connections.

\textbf{Experiments on ImageNet.}
We evaluate the proposed approaches on large-scale dataset ImageNet \cite{mcmahan2016communication}. ImageNet consists of 1.2M training images in 1000 classes. 
The main networks are ResNet-18 and VGG-11. 

\begin{table}
	%	\vspace{-20pt}
	\centering
% 	\captionsetup{labelfont={color=blue},font={color=blue}}
	\caption{Top-1 and Top-5 accuracy by EIF+EMP and baselines with ResNet-18 and VGG-11 on ImageNet.}
	\vspace{-4pt}
	{\begin{tabular}{lllll}
		\toprule 
		Network & Method & Comp. Reduce & \begin{tabular}[c]{@{}l@{}}Accuracy\\ (top-1)\end{tabular} & \begin{tabular}[c]{@{}l@{}}Accuracy\\ (top-5)\end{tabular} \\ \hline
		\multirow{6}{*}{ResNet-18} & SGD (Original) & - & 69.76\% & 89.08\%\\
		& \multirow{2}{*}{\textbf{EIF+EMP}} & \textbf{58.91\%} & \textbf{70.27\%} & \textbf{89.63\%} \\
		&  & \textbf{64.71\%} & \textbf{68.98\%} & \textbf{89.35\%} \\
		& OHEM & 46.67\% & 62.09\% & 87.08\% \\
		& SD & 50.00\% & 65.36\% & 86.41\% \\
		& SMB & 50.00\% & 65.94\% & 87.50\% \\ 
		\hline
		\multirow{5}{*}{VGG-11} & SGD (Original) & - & 70.38\% & 89.81\% \\
		& \multirow{2}{*}{\textbf{EIF+EMP}} & \textbf{51.63\%} & \textbf{70.36\%} & \textbf{89.98\%}\\
		&  & \textbf{60.59\%} & \textbf{70.01\%} & \textbf{89.83\%} \\
		& OHEM & 46.59\% & 56.39\% & 85.62\% \\
		& SMB & 50.00\% & 63.76\% & 86.49\%\\ 
		\bottomrule
	\end{tabular}
	}
	\vspace{-19pt}
	\label{tbl:eifemp_imagenet}
\end{table}

The proposed EIF+EMP effectively reduces the computation cost in training while preserving the accuracy on the large-scale dataset and significantly outperforms the baselines. As shown in Table \ref{tbl:eifemp_imagenet}, when training ResNet-18, with 58.91\% computation reduction in training, EIF+EMP boosts the top-1 accuracy by 0.51\% (70.27\% vs. 69.76\%) and boosts the top-5 accuracy by 0.55\%. With more aggressive computation reduction of 64.71\%, EIF+EMP still boosts the top-5 accuracy by 0.27\% (89.35\% vs. 89.08\%). 
EIF+EMP consistently outperforms the SOTA baselines by a large margin. With larger computation reduction, EIF+EMP achieves higher top-1 accuracy with range [4.33\%, 8.18\%] and higher top-5 accuracy with range [2.13\%, 3.22\%], respectively. SD relies on the residual connections and cannot be applied to VGG-11.
Similar results are observed on VGG-11 as shown in Table \ref{tbl:eifemp_imagenet}.

\vspace{-6pt}
\subsection{Convergence Speed}
\begin{figure}[!htb]
	\centering
	\vspace{-15pt}
	\includegraphics[width=0.9\columnwidth]{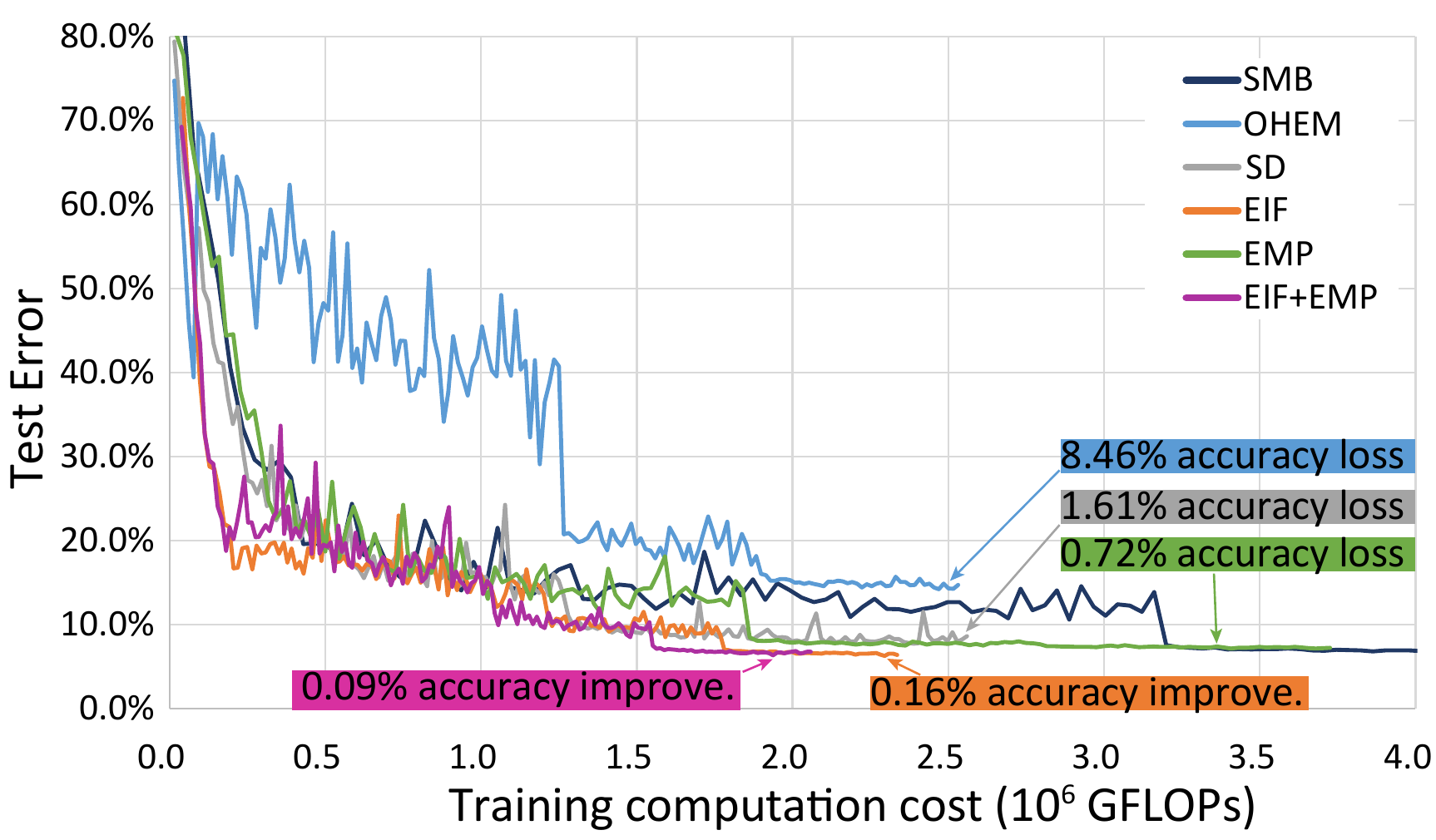}
	\vspace{-4pt}
	\caption{Convergence speed of ResNet-110 on CIFAR-10 during training with different approaches.}
 	\vspace{-12pt}
	\label{fig:converge}
%	\vspace{-8pt}
\end{figure}
The proposed approaches improve the convergence speed in the training process. The test error (i.e. 100\% - accuracy on test dataset) over the computation cost during training is shown in Fig. \ref{fig:converge}. The proposed EIF, EMP and combined EIF+EMP approaches
converge faster than the baselines, represented as lower test error (higher accuracy) with the same computation cost. More specifically, EIF+EMP achieves 3.1x faster convergence and 0.09\% accuracy improvement compared with the standard mini-batch approach (SMB). 
The SOTA baselines OHEM and SD achieve lower convergence speed and larger accuracy loss of 8.46\% and 1.61\%, respectively. 

\vspace{-6pt}
\subsection{Quantitative and Qualitative Analysis}\label{sec:exp_analysis}
\begin{figure}[!htb]
	\centering
	\vspace{-5pt}
	\includegraphics[width=1.0\columnwidth]{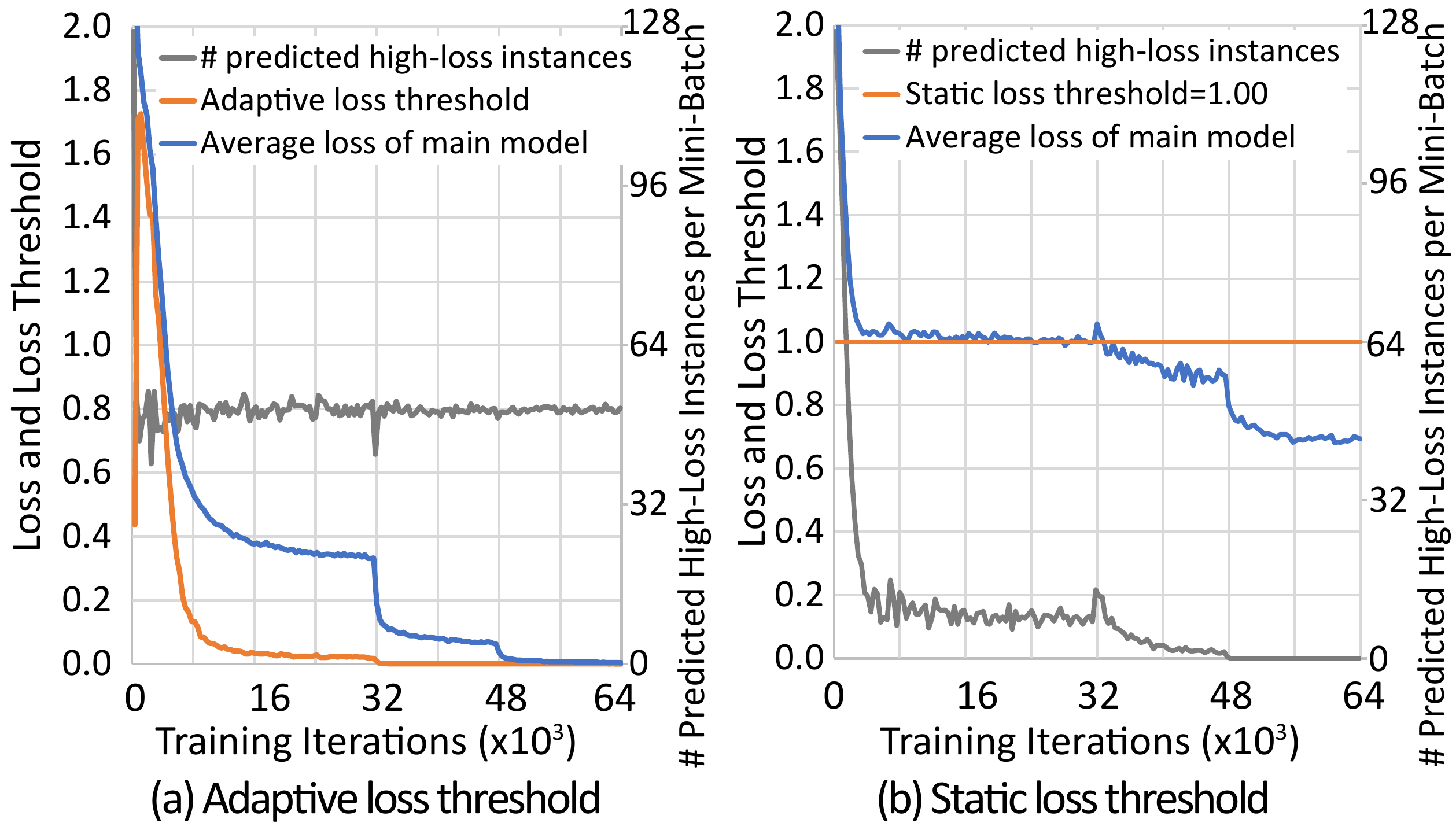}
	\vspace{-15pt}
	\caption{The adaptive loss threshold (left) tracks the state of the main model in the training process and stabilizes the number of preserved instances with predicted high loss by EIF. 
    Static loss threshold (right) cannot generate correct loss labels to train the EIF, which results in incorrect number predicted high-loss instances and a high average loss of the main model.}
	\vspace{-10pt}
	\label{fig:loss_threshold}
%	\vspace{-8pt}
\end{figure}
\textbf{Effectiveness of Adaptive Loss Threshold.}
The proposed early instance filter effectively predicts a pre-defined percentage of input instances as high-loss and the adaptive loss threshold effectively adjusts the loss threshold as the labeling strategy to train the filter. 
In Fig. \ref{fig:loss_threshold}(a), the pre-defined high loss ratio is 40\% for training ResNet-110 on CIFAR-10.
The number of predicted high-loss instances, averaged every 390 iterations, is stabilized at about 51, which effectively selects 40\% high-loss instances on average from 128 instances in each mini-batch.
As the average loss decreases, the adaptive loss threshold also decreases following a similar pattern to closely track the latest state of the main network.

We further compare the proposed adaptive loss threshold with the static loss threshold. With a static loss threshold 1.0, the number of predicted high-loss instances per mini-batch and the average loss of the main model in the training process are shown in Fig. \ref{fig:loss_threshold}(b).
The goal of training is to minimize the loss of the main model. However, the static loss threshold cannot effectively decrease the loss of the main model as shown in the blue line, and results in low accuracy.
This is because the static loss threshold cannot track the latest state of the main model. Therefore, it cannot effectively stabilize the number of predicted high-loss instances to train the main model.
The static loss threshold only achieves 80.83\% final accuracy of the main model. 
Different from this, the proposed adaptive loss threshold can effectively minimize the loss of the main model
and achieves high accuracy of 94.24\%.

\begin{figure}[!htb]
	\centering
	\vspace{-12pt}
	\includegraphics[width=0.8\columnwidth]{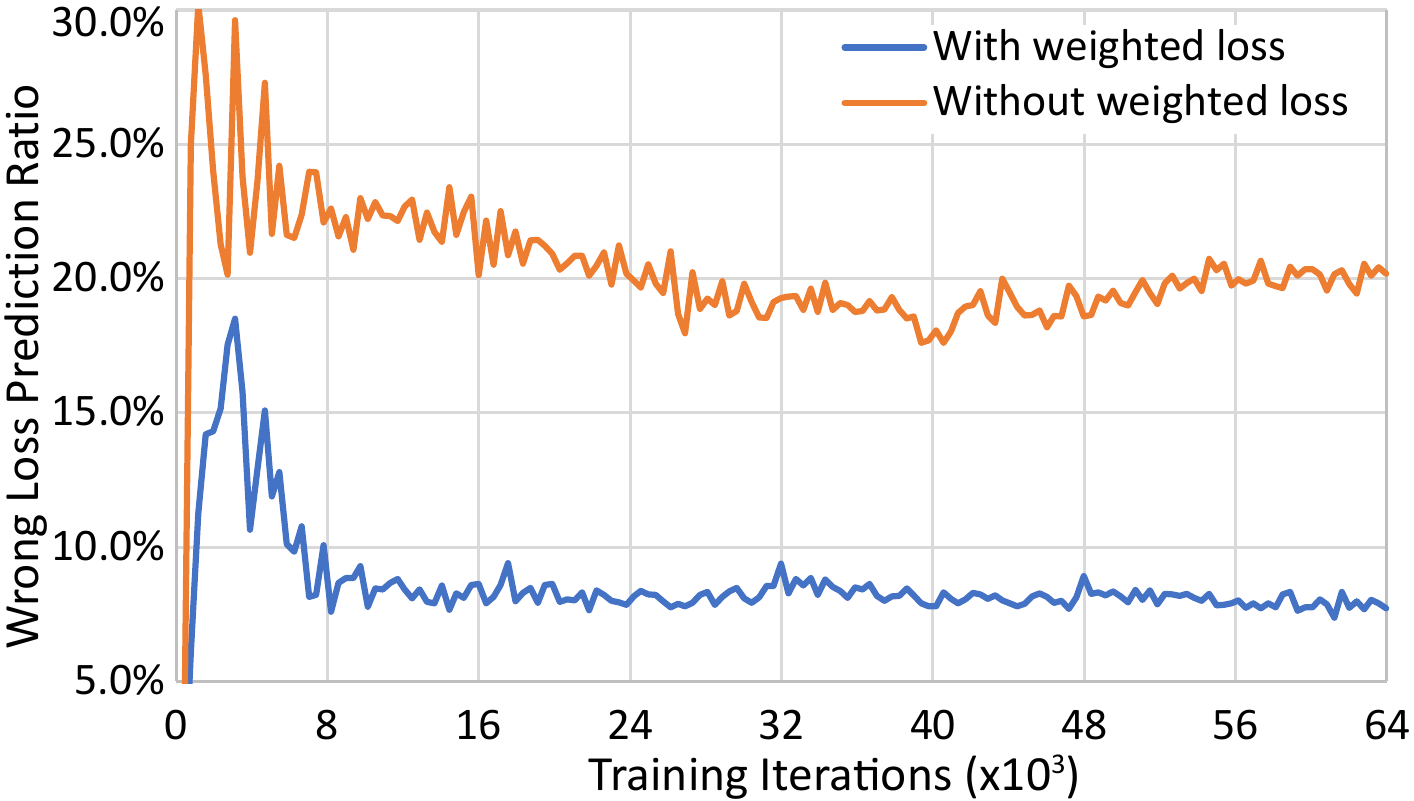}
	\vspace{-6pt}
% 	\captionsetup{labelfont={color=blue},font={color=blue}}
	\caption{Incorrect loss prediction ratio of EIF with and without weighted loss.}
	\vspace{-8pt}
	\label{fig:exp_weighted_loss}
%	\vspace{-8pt}
\end{figure}

\textbf{Effectiveness of Weighted Loss for Training EIF.}
The weighted loss in Eq.(\ref{equ:loss_filter}) effectively trains the EIF network to make accurate loss prediction, which eventually results in higher accuracy of the main model. 
As shown in Fig. \ref{fig:exp_weighted_loss}, when the weighted loss is employed, the wrong loss prediction ratio by the EIF is much lower than that without weighted loss. The pre-defined high-loss ratio is 30\%, and the corresponding low-loss ratio is 70\%. This high-loss ratio makes the number of high-loss and low-loss instances unbalanced in the input stream.

When the weighted loss is used for training EIF, the average wrong loss prediction ratio by EIF is reduced from 20.31\% to 8.59\%. 
This accurate loss prediction effectively selects high-loss instances to train the main model and results in significantly higher accuracy of the main model, which is 94.05\% with weighted loss vs. 90.58\% without weighted loss.

\begin{figure}[!htb]
	\centering
	\vspace{-12pt}
    \includegraphics[width=1.0\columnwidth]{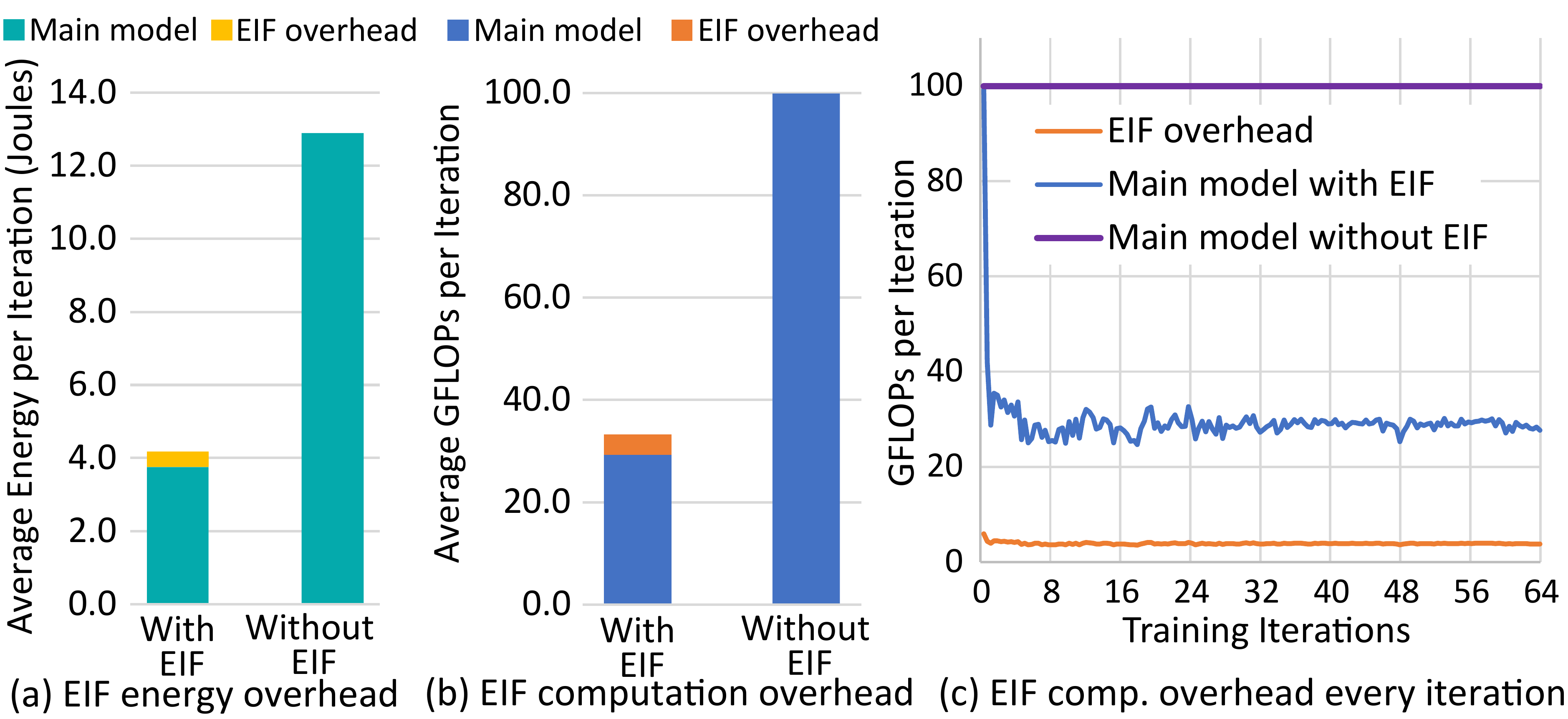}
	\vspace{-12pt}
% 	\captionsetup{labelfont={color=blue},font={color=blue}}
	\caption{Energy and computation overhead of EIF. Energy overhead is measured on NVIDIA Jetson TX2 mobile GPU.}
	\vspace{-10pt}
	\label{fig:EIF_overhead}
%	\vspace{-8pt}
\end{figure}

\textbf{Overhead of EIF.}
The proposed early instance filter has marginal energy and computation overhead. 
The average energy and computation overhead of the EIF network per training iteration (e.g. one mini-batch of 128 instances) when training ResNet-110 on CIFAR-10 dataset is shown in Fig. \ref{fig:EIF_overhead}. 
As shown in the yellow bar in Fig. \ref{fig:EIF_overhead}(a), the energy overhead of the EIF network (measured on NVIDIA Jetson TX2) is 0.43J per iteration, which is 10.22\% of the total energy cost 4.18J when training with EIF. Without EIF, the energy cost is 12.90J per iteration.
As shown in Fig. \ref{fig:EIF_overhead}(b), the computation overhead of EIF is 3.88 GFLOPs, which is 11.65\% of the total computation cost 33.21 GFLOPs when training with EIF. Without EIF, the computation cost is 99.91 GFLOPs per iteration.
The detailed EIF computation overhead across all training iterations are shown in Fig. \ref{fig:EIF_overhead}(c). 
While the overhead of EIF is not zero, the proposed approach achieves 67.60\% energy saving and 66.76\% computation saving while fully preserving the accuracy.

\begin{figure}[!htb]
	\centering
	\vspace{-10pt}
	\includegraphics[width=0.9\columnwidth]{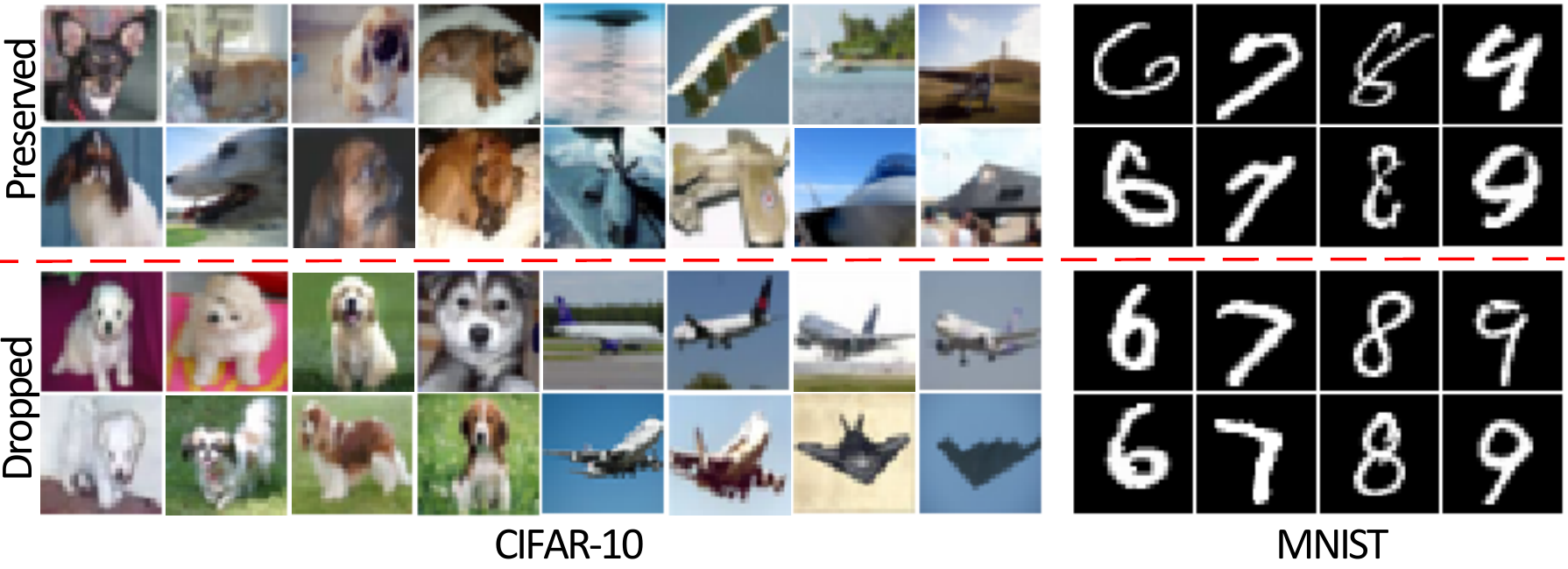}
	\vspace{-6pt}
	\caption{Preserved and dropped instances by EIF when training ResNet-110 on CIFAR-10 and LeNet on MINST.}
%	\vspace{-10pt}
	\label{fig:visual_cifar_mnist}
	\vspace{-8pt}
\end{figure}
\textbf{Preserved and Dropped Instances by EIF.}
To better understand the instances selected by the early instance filter, we cluster the instances that the filter preserves and drops when training ResNet-110 on CIFAR-10 and LeNet on MNIST, as shown in Fig. \ref{fig:visual_cifar_mnist}. 
We find that the dropped instances show the full objects with typical characteristics. The preserved instances either only show part of the object or show non-typical characteristics, even hard for humans to understand. This result shows the early instance filter can effectively find important instances to train the network.

\begin{figure}[!htb]
	\centering
	\vspace{-14pt}
	\includegraphics[width=1.0\columnwidth]{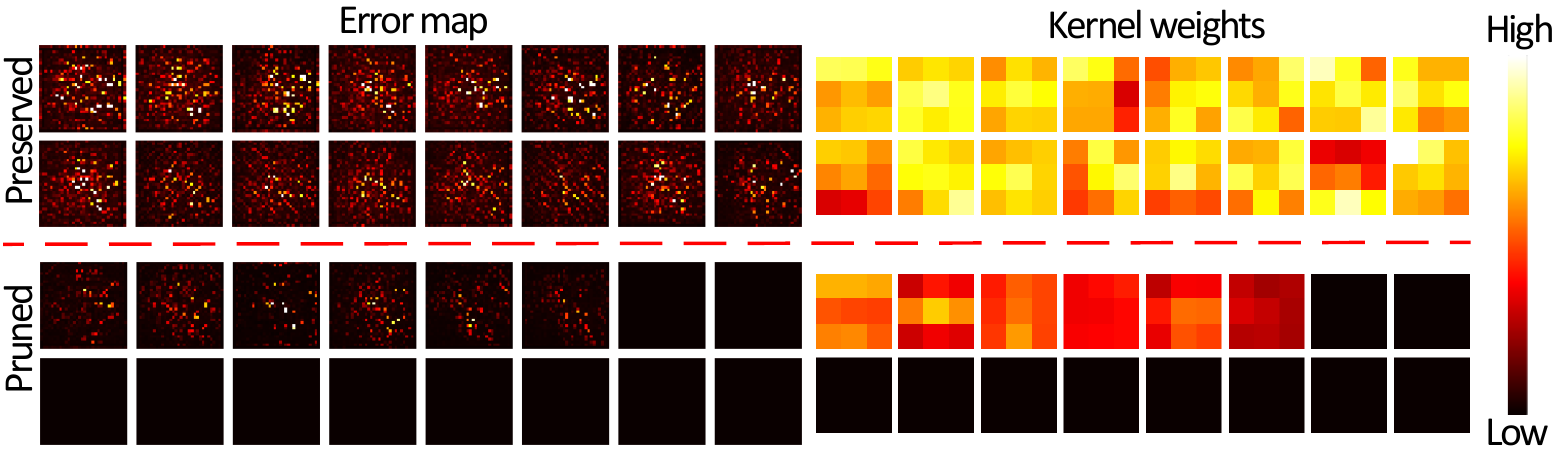}
	\vspace{-18pt}
	\caption{Visualization of the pruned and preserved channels in error map and corresponding convolutional kernel weights.}
	\vspace{-2pt}
	\label{fig:visual_error_map}
	\vspace{-8pt}
\end{figure}
\textbf{Analysis of Error Map Pruning.}
To better understand the pruned and preserved channels in the backward pass by error map pruning, we visualize them to analyze the effectiveness of the proposed channel selection approach. The preserved and pruned channels in the error map and corresponding kernel weights in conv2 layer of VGG-16 are shown in Fig. \ref{fig:visual_error_map}. The 16 channels with the highest/lowest proposed importance scores are shown on the top left and bottom left, respectively, and their corresponding convolutional kernel weights are shown on the right. The pruned channels are darker with smaller values than the preserved channels, which are brighter with larger values. Similarly, the kernel weights corresponding to the pruned channels have smaller values than the preserved ones. Therefore, the pruned channels will have the least influence on both the error propagation and computation and weight gradients. This result shows the proposed error map pruning approach effectively selects the channels to prune to minimize the influence on training.

\vspace{-10pt}
\subsection{Practical Energy Saving on Hardware Platforms}\label{sec:hw_exp}
The energy cost of training consists of both the computation cost and the memory access cost. 
While the former one dominates the energy cost and is represented by the commonly used metric FLOPs \cite{sandler2018mobilenetv2}, 
the \emph{energy saving} ratio can be slightly different from the \emph{computation reduction} ratio.
To evaluate the practical energy saving, we conduct extensive experiments on two edge platforms and evaluate the proposed approaches in terms of \emph{practical energy saving} and \emph{accuracy}.

\begin{figure}[!htb]
	\centering
	\vspace{-12pt}
	\includegraphics[width=0.9\columnwidth]{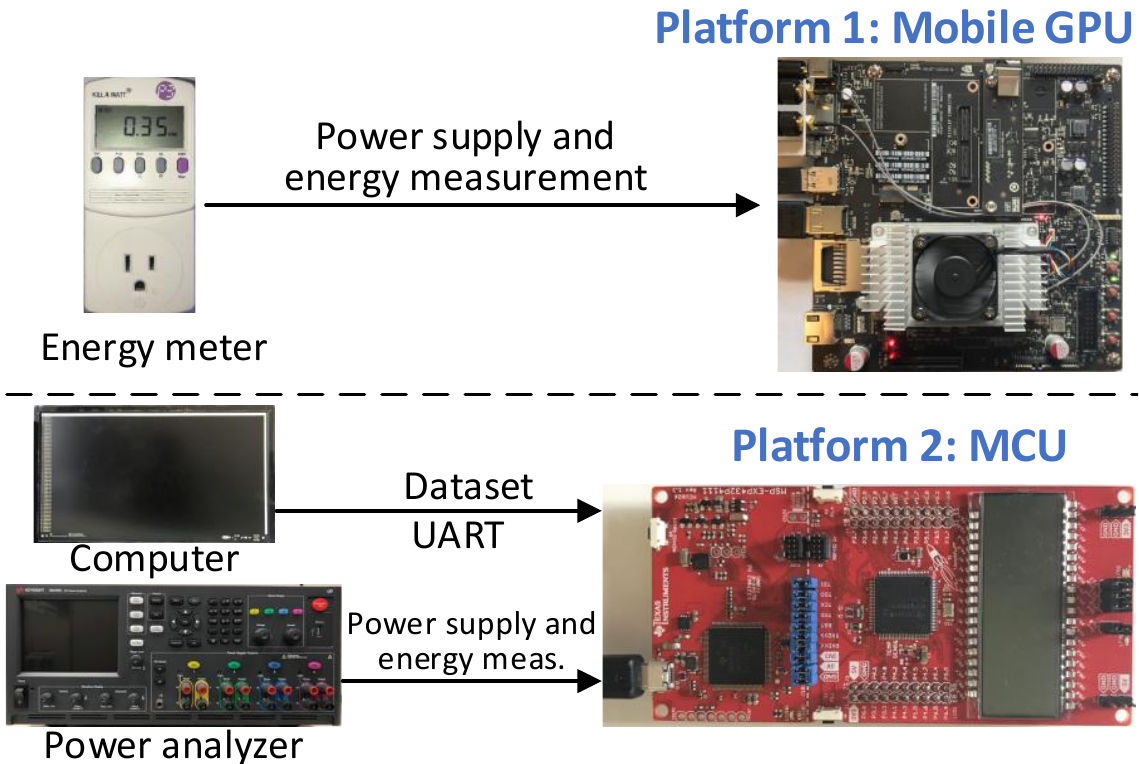}
	\vspace{-6pt}
% 	\captionsetup{labelfont={color=blue},font={color=blue}}
	\caption{Energy measurement setup for training on two edge platforms, including mobile-level devices (top) and sensor node-level devices (bottom).}
	\vspace{-6pt}
	\label{fig:hardware}
%	\vspace{-8pt}
\end{figure}

\textbf{Hardware Setup.}
We apply the proposed training approach on two edge platforms to evaluate \textit{realistic energy saving}. 
\textbf{For mobile-level devices,} we train ResNet-110, ResNet-74, and VGG-16 on an NVIDIA Jetson TX2 mobile GPU \cite{jetsontx2} with CIFAR-10 and CIFAR-100 datasets by PyTorch 1.1. 
We use an energy meter to measure the energy cost as shown on the top of Fig. \ref{fig:hardware}.
\textbf{For sensor node-level devices,} we train LeNet on the MSP432 MCU \cite{msp432}.
We use C language to implement the training process on MCU. Since the MCU cannot store the entire dataset, we use a computer to feed the training data into the MCU via UART in the training process. We use the Keysight N6705C power analyzer to measure the energy cost on the MCU as shown on the bottom of Fig. \ref{fig:hardware}.

\begin{figure}[!htb]
	\centering
    \vspace{-6pt}
    \includegraphics[width=1.0\columnwidth]{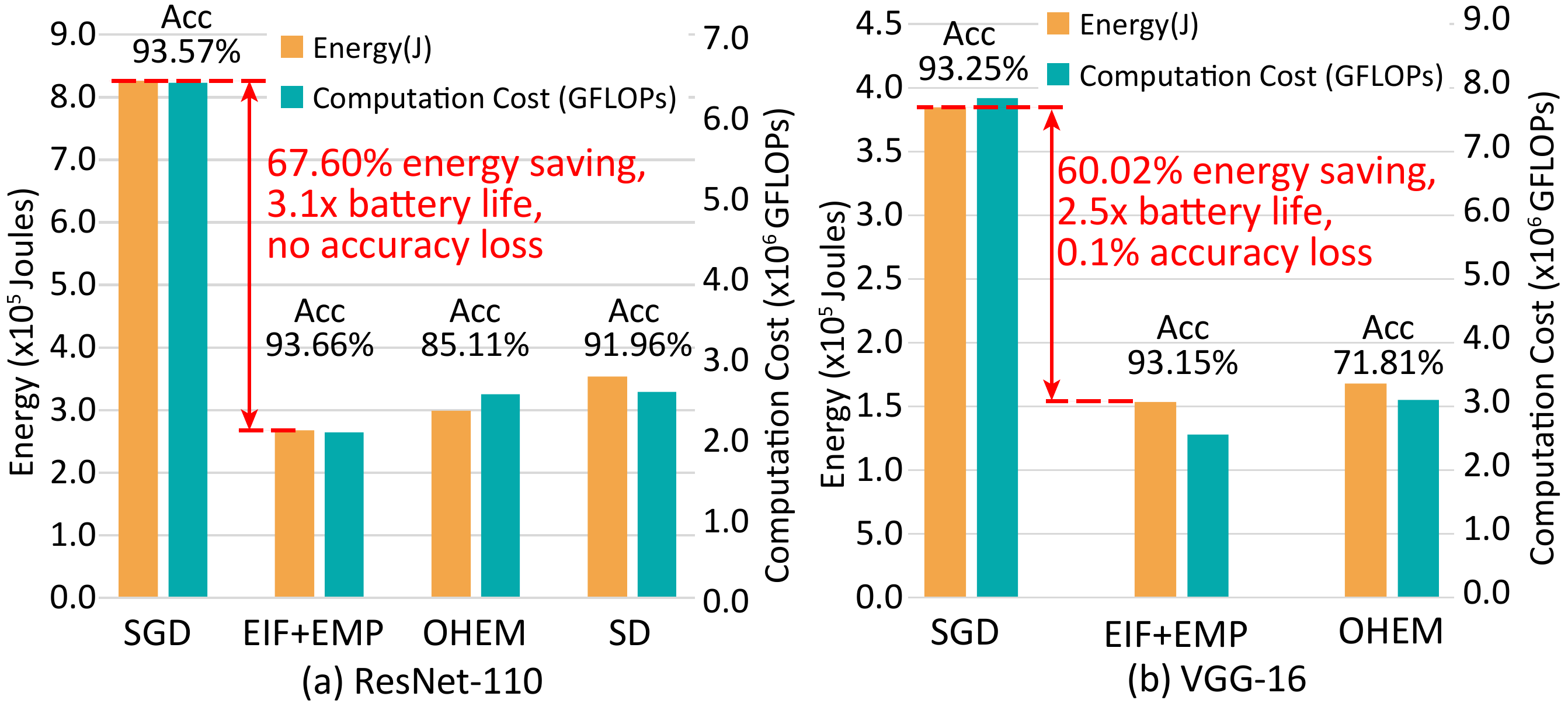}
	\vspace{-20pt}
	\caption{Energy saving when training ResNet-110 and VGG-16 on Nvidia Jetson TX2 \cite{jetsontx2} mobile GPU with CIFAR-10 dataset. EIF+EMP prolongs 2.5x to 3.1x battery life without any or with marginal accuracy loss.}
	\vspace{-10pt}
	\label{fig:exp9_energy_mobile}
%	\vspace{-8pt}
\end{figure}

\textbf{Energy Saving of Training on Mobile GPU.}
We evaluate the energy saving by EIF+EMP on mobile-level devices. We repeat all the experiments in Table \ref{tbl:res110_quant} and \ref{tbl:eifemp_cifar100} on the mobile GPU to measure the practical energy saving, except for the LeNet, which will be evaluated on MCU. 
Our approach effectively reduces the energy cost of on-device training. Compared with the original SGD, the proposed EIF+EMP achieves energy saving of 67.60\%, 63.57\%, and 60.02\% in the training of ResNet-110, ResNet-74, and VGG-16 on CIFAR-10, respectively, as shown in Fig. \ref{fig:exp9_energy_mobile} (the result of ResNet-74 is not shown for conciseness). The energy savings prolong battery life by 3.1x, 2.7x, and 2.5x while improving the accuracy or incurring a slight 0.1\% accuracy loss. 
Compared with the SOTA baselines OHEM and SD, our approach achieves significantly higher accuracy when similar energy saving is achieved.
SD relies on the residual connections and cannot be applied to VGG-16.
Besides, the practical energy saving ratios are very close to the computation reduction ratios represented by FLOPs, which shows the computation reduction in FLOPs can generalize well to energy saving on hardware platforms.
Similar results are observed on the CIFAR-100, on which we achieve 54.22\% and 46.64\% energy saving (2.2x and 1.9x battery life) for ResNet-110 and VGG-16 without any accuracy loss, respectively.

\begin{figure}[!htb]
	\centering
	\vspace{-15pt}
	\includegraphics[width=0.7\columnwidth]{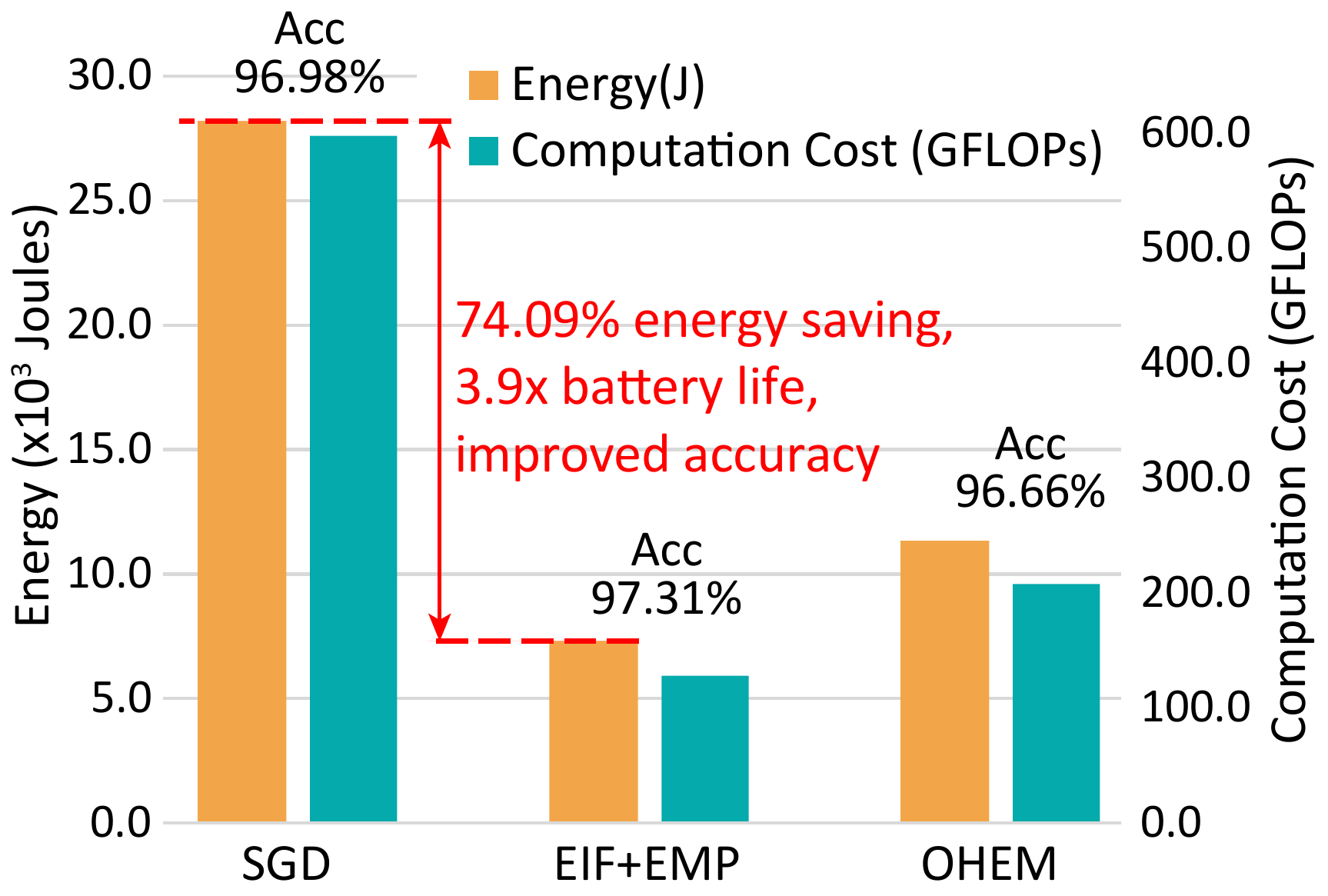}
	\vspace{-6pt}
	\caption{Energy saving when training LeNet on MSP432 \cite{msp432} MCU. EIF+EMP prolongs 3.9x battery life.}
	\vspace{-10pt}
	\label{fig:exp_mcu_lenet}
\end{figure}

\textbf{Energy Saving of Training on MCU.}
We evaluate the energy saving by EIF+EMP on sensor node-level devices (i.e. MCUs). We train LeNet on MCU MSP432 for one epoch including 60000 instances and measure the energy cost and accuracy.
Due to the limited runtime memory, we set the batch size to 1. 
Since the original SGD approach takes too long (i.e. about 50 days) to complete on MCU, we conduct 10\% training iterations of one epoch on MCU and estimate the total energy cost by multiplying the measured energy cost by 10. The accuracy of the original SGD is measured on the P100 GPU after finishing one epoch.
OHEM cannot be applied to MCUs because it needs batch-wise loss values for instance selection. To compare with OHEM, we measure its energy cost on MCU by completing its computation and ignoring the accuracy.
The accuracy of OHEM is evaluated on P100 GPU.

EIF+EMP significantly reduces the energy cost of training on MCUs and effectively prolongs battery life. As shown in Fig. \ref{fig:exp_mcu_lenet}, when training LeNet on MSP432 MCU, EIF+EMP effectively reduces the energy cost by 74.09\% while improving the accuracy by 0.33\%. 
This prolongs battery life by 3.9x. 
OHEM, while not fully feasible on MCU, achieves much lower energy saving of 59.78\% with an accuracy loss of 0.32\%.
This result shows EIF+EMP greatly improves the battery life of tiny sensor nodes and outperforms the baselines.

% \vspace{-10pt}
\section{Conclusion}
This work aims to enable on-device training of convolutional neural networks by reducing the computation cost at training time. We propose two complementary approaches to reduce the computation cost: early instance filtering (EIF), which selects important instances for training the network and drops trivial ones, and error map pruning (EMP), which prunes insignificant channels in error map in back-propagation. Experimental results show superior computation reduction with higher accuracy compared with state-of-the-art techniques.

\bibliographystyle{IEEEtran}
\bibliography{IEEEabrv,IEEEexample}
\end{document}